\documentclass[11pt]{article}

\usepackage[table]{xcolor}
\usepackage[preprint]{acl}

\usepackage[T1]{fontenc}
\usepackage[utf8]{inputenc}
\usepackage{times}
\usepackage{latexsym}
\usepackage{microtype}
\usepackage{inconsolata}
\usepackage{xurl}

\usepackage{amsmath}
\usepackage{amssymb}

\usepackage{graphicx}
\usepackage{float}
\usepackage{booktabs}
\usepackage{multirow}
\usepackage{tabularx}
\usepackage{array}
\usepackage{arydshln}
\usepackage{caption}

\usepackage[most]{tcolorbox}
\tcbuselibrary{listings}
\usepackage{enumitem}

\newcolumntype{Y}{>{\centering\arraybackslash}X}
\newcolumntype{C}[1]{>{\centering\arraybackslash}m{#1}}
\newcolumntype{L}[1]{>{\raggedright\arraybackslash}m{#1}}

\definecolor{mywheat}{RGB}{255,235,205}
\definecolor{mylightblue}{RGB}{221,235,247}

\newcommand{\best}[1]{\colorbox{mywheat}{{\bfseries\boldmath #1}}}
\newcommand{\second}[1]{\colorbox{mylightblue}{#1}}

\definecolor{airlineblue}{HTML}{1F5FA8}
\definecolor{hotelgold}{HTML}{B7791F}
\definecolor{restaurantorange}{HTML}{C05621}
\definecolor{retailgreen}{HTML}{2F855A}
\definecolor{platformpurple}{HTML}{6B46C1}

\definecolor{journeylight}{HTML}{F7FAFC}
\definecolor{journeygray}{HTML}{5B6472}

\colorlet{journeycolor}{airlineblue}
\colorlet{journeydark}{airlineblue}

\newcommand{\setjourneycolor}[1]{%
  \colorlet{journeycolor}{#1}%
  \colorlet{journeydark}{#1}%
}

\newtcolorbox{journeymapbox}[1]{
  enhanced,
  colback=journeylight,
  colframe=journeycolor,
  boxrule=0.55pt,
  arc=1.6mm,
  left=5pt,
  right=5pt,
  top=5pt,
  bottom=5pt,
  title=\textbf{#1},
  colbacktitle=journeycolor,
  coltitle=white,
  fonttitle=\small,
  boxed title style={
    colback=journeycolor,
    colframe=journeycolor,
    boxrule=0pt,
    arc=1.2mm,
    left=4pt,
    right=4pt,
    top=1pt,
    bottom=1pt
  }
}

\newcommand{\phaseblock}[3]{%
  \vspace{4pt}
  \noindent
  {\colorbox{journeycolor}{\textcolor{white}{\tiny\bfseries\strut #1}}}
  \hspace{3pt}
  {\scriptsize\bfseries\textcolor{journeydark}{#2}}\par
  \vspace{0.5pt}
  {\tiny\itshape\textcolor{journeygray}{#3}}\par
  \vspace{2pt}
}

\newcommand{\maincat}[2]{%
  \vspace{3pt}
  \noindent\hspace{0.9em}%
  {\tiny\bfseries\textcolor{journeycolor}{#1}~#2}\par
  \vspace{1pt}
}

\newenvironment{compactsubcats}{
  \begin{itemize}[
    leftmargin=3.4em,
    labelsep=0.45em,
    itemsep=1.2pt,
    topsep=1.5pt,
    parsep=0pt,
    partopsep=0pt
  ]
  \tiny
}{
  \end{itemize}
  \vspace{2pt}
}

\newcommand{\subcat}[2]{%
  \item[\textcolor{journeycolor}{#1}] #2
}

\definecolor{promptgray}{HTML}{4A5568}
\definecolor{promptlight}{HTML}{F7FAFC}
\definecolor{promptborder}{HTML}{A0AEC0}

\newtcblisting{promptlisting}[1]{
  enhanced,
  listing only,
  listing options={
    basicstyle=\ttfamily\scriptsize,
    breaklines=true,
    breakatwhitespace=true,
    breakautoindent=false,
    breakindent=0pt,
    columns=fullflexible,
    keepspaces=true,
    showstringspaces=false,
    tabsize=2
  },
  colback=promptlight,
  colframe=promptborder,
  boxrule=0.55pt,
  arc=1.6mm,
  left=6pt,
  right=6pt,
  top=6pt,
  bottom=6pt,
  title=\textbf{#1},
  colbacktitle=promptgray,
  coltitle=white,
  fonttitle=\small,
  boxed title style={
    colback=promptgray,
    colframe=promptgray,
    boxrule=0pt,
    arc=1.2mm,
    left=4pt,
    right=4pt,
    top=1pt,
    bottom=1pt
  }
}

\definecolor{caseblue}{HTML}{5B728A}
\definecolor{casebluegray}{HTML}{4F6F8F}
\definecolor{caselight}{HTML}{F8FAFC}
\definecolor{caseborder}{HTML}{B8C2CC}
\definecolor{turnlight}{HTML}{FFFFFF}
\definecolor{turnhead}{HTML}{EAF1F8}
\definecolor{metaback}{HTML}{F1F5F9}

\newtcolorbox{casebox}[1]{
  enhanced,
  colback=caselight,
  colframe=caseborder,
  boxrule=0.55pt,
  arc=1.6mm,
  left=6pt,
  right=6pt,
  top=6pt,
  bottom=6pt,
  title=\textbf{#1},
  colbacktitle=caseblue,
  coltitle=white,
  fonttitle=\small,
  boxed title style={
    colback=caseblue,
    colframe=caseblue,
    boxrule=0pt,
    arc=1.2mm,
    left=4pt,
    right=4pt,
    top=1pt,
    bottom=1pt
  }
}

\newtcolorbox{casebackground}{
  enhanced,
  colback=metaback,
  colframe=caseborder,
  boxrule=0.35pt,
  arc=1.2mm,
  left=5pt,
  right=5pt,
  top=4pt,
  bottom=4pt
}

\newtcolorbox{dialogueturn}[1]{
  enhanced,
  colback=turnlight,
  colframe=caseborder,
  boxrule=0.35pt,
  arc=1.2mm,
  left=5pt,
  right=5pt,
  top=4pt,
  bottom=4pt,
  title=\textbf{#1},
  colbacktitle=turnhead,
  coltitle=casebluegray,
  fonttitle=\scriptsize,
  boxed title style={
    colback=turnhead,
    colframe=turnhead,
    boxrule=0pt,
    arc=1.0mm,
    left=3pt,
    right=3pt,
    top=0.5pt,
    bottom=0.5pt
  }
}

\title{Can LLM Agents Resolve Service Failures? Simulating Multi-Turn Complaint Handling under Information Asymmetry}

\author{
  \textbf{Wailing Ng} \quad
  \textbf{Mengze Hong} \quad
  \textbf{Chen Jason Zhang} \quad
  \textbf{Shiqi Fan} \\
  The Hong Kong Polytechnic University \\
}

\begin{document}
\maketitle
\begin{abstract}

LLM agents are widely adopted in service chatbots for rule-guided support and information access, yet emotionally demanding complaint handling still relies heavily on human agents. We introduce \textbf{\textsc{ServiceBench}}, to our knowledge the first benchmark dedicated to complaint handling, with 2,500 scenarios spanning 184 complaint categories and five service industries. We propose \textbf{Asymmetric Progressive Complaint Simulation (APCS)} to simulate realistic complaint handling, together with a justice-based evaluation framework. Our experiments reveal that even capable LLM agents struggle to handle complaints effectively. They lack the ability to proactively gather information and strategically guide conversations toward resolution, often resorting to premature compensation. We further show that targeted post-training with SFT and DPO substantially improves these capabilities, enabling a smaller model to match or surpass much larger models on several outcomes.

\end{abstract}

\section{Introduction}

Complaint handling is a central component of service recovery, through which service providers respond to failures and attempt to restore damaged customer relationships \cite{spreng1995service,smith1999model}. Poor complaint handling can deepen dissatisfaction and increase the cost of service recovery \cite{bell1987service,meyer2023relationship}. Although LLM-based agents are increasingly used for routine customer support, complaint handling remains a challenging setting in which effective service still depends heavily on human judgment \cite{tan2025chatgpt}.

\begin{figure}[t]
    \centering
    \includegraphics[width=0.95\columnwidth]{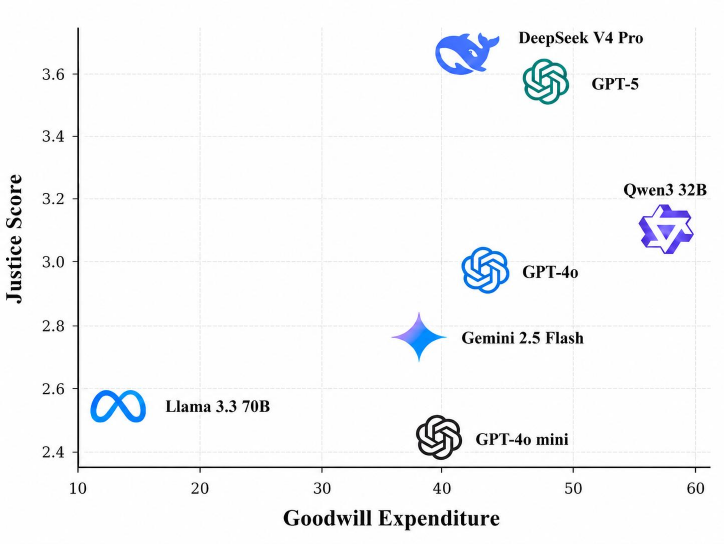}
    \caption{Comparison of complaint handling quality and goodwill expenditure across LLM agents.}
    \label{fig:goodwill_vs_justice}
\end{figure}

Existing customer service benchmarks primarily evaluate operational capabilities such as domain knowledge \cite{li2025performance,wang2025ecom,lee2024telbench,wang2025teleeval}, tool use \cite{yao2024taubench,shi2024chops,huang2025crmarena}, and workflow compliance \cite{nandi2025sop,balaji2026beyond,barres2025evaluating}. These benchmarks capture important operational capabilities, but they do not evaluate how agents handle complaints after service failures. As LLMs are increasingly deployed in customer service \cite{su2025llm, wang2025ecom, hong2025dial, hong2025qualbench}, evaluating their ability to handle complaints becomes critical.

To address this gap, we introduce \textbf{\textsc{ServiceBench}}, a multi-turn benchmark containing 2,500 scenarios across 184 complaint categories and five service industries. We construct the benchmark from real complaint dialogues using an industry specific customer journey guided pipeline. This pipeline combines data driven topic discovery with expert guided coding to organize service failures by journey stage, providing broad and structured coverage of real world complaint contexts.

Complaint handling often begins before the service agent has access to the full circumstances of a complaint. To evaluate agents under this setting, we propose \textbf{Asymmetric Progressive Complaint Simulation (APCS)}, which controls information exposure within a shared complaint scenario and conditions subsequent customer disclosure and resolution acceptance on the agent's responses. Since resolution acceptance alone does not capture handling quality, we further evaluate completed interactions using a framework grounded in justice theory \cite{smith1999model}, covering interactional, procedural, and distributive justice.

Our extensive experiments show that DeepSeek-V4-Pro and GPT-5 achieve the strongest overall complaint handling performance (Figure~\ref{fig:goodwill_vs_justice}). By examining the timing of goodwill offers and the information available at that point, we find that agents often offer compensation before sufficiently understanding customer complaints. When provided with complete customer information from the outset, agents perform substantially better. These findings suggest that current agents struggle to proactively gather information and guide conversations toward resolution, often using compensation as a shortcut. Finally, we show that targeted training on these behaviors through SFT and DPO substantially improves complaint handling, enabling Qwen3-8B to match or surpass much larger general purpose models on several evaluation outcomes.

Our contributions are threefold:

\begin{itemize}
    \item We introduce \textbf{\textsc{ServiceBench}}, to our knowledge the first multi-turn benchmark designed for complaint handling spanning 184 complaint categories and 5 service industries.

    \item We propose \textbf{Asymmetric Progressive Complaint Simulation (APCS)} to model realistic complaint interactions and a justice-based evaluation framework to evaluate agent capability in complaint handling. 

    \item Extensive experiments reveal systematic behavioral limitations in current models and further show that targeted post-training can substantially mitigate these limitations.
\end{itemize}

\section{Related work}

\begin{figure*}[t]
    \centering
    \includegraphics[width=\textwidth]{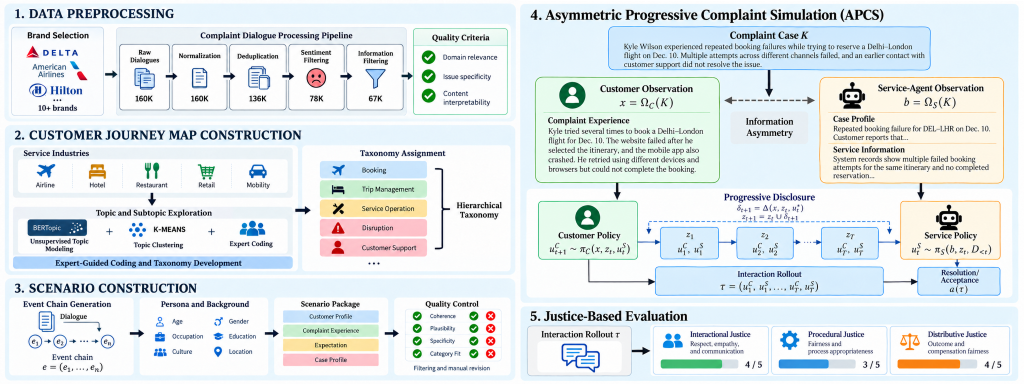}
    \caption{Overall framework of \textsc{ServiceBench}. We construct scenarios from real complaint dialogues on X and expert guided industry level customer journey maps to ensure realistic, diverse, and operationally grounded service failure contexts. We then apply Asymmetric Progressive Complaint Simulation (APCS) to generate multi-turn complaint handling interactions under information asymmetry. Finally, an LLM-based judge evaluates the completed interactions based on interactional, procedural, and distributive justice.}
    \label{fig:servicebench_overview}
\end{figure*}

\subsection{Service Recovery and Justice in Complaint Handling}

Service recovery concerns how service providers respond to failures and restore customer evaluations and relationships \cite{spreng1995service}. Prior research distinguishes symbolic recovery, which relies on interpersonal and communicative responses, from utilitarian recovery, which provides tangible remedies \cite{bell1987service,smith1999model,meyer2023relationship}. Effective complaint handling requires more than selecting a remedy, as customers also evaluate how they are treated and how the resolution process is conducted. Justice theory formalizes these complementary aspects through interactional, procedural, and distributive justice \cite{smith1999model}. This perspective provides a principled basis for evaluating complaint handling beyond task completion.

\subsection{Benchmarks for Customer Service Agents}

There is growing interest in benchmarking LLMs for domain specific customer service applications. Existing work primarily targets three forms of operational capability. Domain knowledge benchmarks assess service specific knowledge \cite{li2025performance,lee2024telbench,wang2025teleeval,wang2025ecom}. Tool use benchmarks evaluate interaction with external systems and service task execution \cite{yao2024taubench,barres2025evaluating,huang2025crmarena}. Workflow compliance benchmarks examine adherence to prescribed procedures and policies \cite{nandi2025sop,balaji2026beyond}. Their evaluation, however, remains largely centered on operational correctness. Complaint handling introduces a different evaluation target, as service quality depends not only on what resolution is reached but also on how the customer is treated throughout the process.

\subsection{User Simulation for Interactive Agents}

Simulated users provide a scalable mechanism for constructing dynamic multi-turn interactions with conversational agents \cite{dou2025simulatorarena,kwon2025evaluating,wu2025socialcc,shi2025educationq}. Recent work has used simulation to evaluate task specific interaction capabilities that are difficult to capture through static benchmarks, including conflict dialogue \cite{kwon2025evaluating}, cultural competence \cite{wu2025socialcc}, and teaching scenarios \cite{shi2025educationq}. Studies have also examined the behavioral fidelity and reliability of simulated users as proxies for human interaction \cite{dou2025simulatorarena}. In this work, we design a simulation framework for complaint handling to evaluate agent capabilities.

\section{Methodology}
\subsection{Task Formulation}
\label{sec:task-formulation}

We formulate complaint handling as a sequential decision problem. Following a service failure, a customer $C$ initiates a complaint with a service agent $S$. At turn $t$, the agent conditions its decision on the current service context $c_t$ and interaction history $D_{<t}$ and produces a response $u_t^S$ according to a service policy $\pi_S$. Each response influences the customer's subsequent behavior $u_{t+1}^C$, and the interaction produces a rollout $\tau=(u_1^C,u_1^S,\ldots,u_T^C,u_T^S)$. Let $Q(\tau)$ denote the resolution quality of the rollout. The objective is to select a sequence of service responses that progressively resolves the complaint and leads the interaction toward an appropriate resolution.

\begin{equation}
\pi_S^{*}
=
\arg\max_{\pi_S}
\mathbb{E}_{\tau \sim \pi_S}
\left[
Q(\tau)
\right].
\end{equation}

\subsection{Benchmark Construction}
\label{sec:benchmark-construction}

\paragraph{Complaint Coverage and Taxonomy.}

\textsc{ServiceBench} provides broad coverage of real world complaint handling, with 2,500 scenarios spanning 184 complaint categories and 37 customer journey stages across 5 service industries: airlines, hotels, restaurants, retail, and ride hailing and delivery. The benchmark is grounded in public complaint dialogues between customers and brand customer service accounts on X \cite{stuart_axelbrooke_2017}, providing empirical grounding in real service failures and customer dissatisfaction. To organize these heterogeneous complaints systematically, we construct industry specific customer journey maps \cite{rosenbaum2017create} through data driven topic discovery and expert guided coding. This process yields an interpretable taxonomy aligned with service stages and touchpoints, enabling balanced coverage of diverse complaint contexts. Detailed benchmark information are provided in Appendix~\ref{app:benchmark-construction}.

\paragraph{Case Reconstruction.} 
 
Real complaint dialogues provide strong factual grounding but are often too sparse to support realistic interactive evaluation. We therefore develop a reconstruction pipeline that transforms each complaint into a structured complaint case while preserving the factual anchors of the source dialogue. We first identify the core complaint facts and reconstruct a plausible sequence of service events that organizes the key events leading to the reported failure and provides a structured backbone for subsequent case generation. Based on this event sequence, we construct the customer experience background and resolution expectation. A company case record is then inferred using a manually designed industry specific schema.

\paragraph{Quality Control.}  
  
We construct 14,373 candidate complaint cases and filter them based on coherence, plausibility, specificity, and category fit. Overall, 86.68\% of candidates meet the filtering criteria. From this pool, we manually select 2,500 scenarios that best match realistic service situations while preserving the complaint information observed in the raw dialogues. We then refine the selected cases to further improve realism and internal consistency before inclusion in the final benchmark.

\subsection{Asymmetric Progressive Complaint Simulation}
\label{sec:simulation}

Complaint handling often begins before the service agent has access to the full circumstances of the complaint. This information asymmetry creates a challenging setting for service resolution. To capture this setting, we introduce \textbf{Asymmetric Progressive Complaint Simulation (APCS)} to enable more realistic evaluation of complaint handling.

\paragraph{Asymmetric Observations.}

Complaint handling involves a fundamental information asymmetry between the customer and service agent. APCS models each complaint with an underlying case representation $\mathcal{K}$ and derives role conditioned observations through separate observation mappings. The customer observation captures the complete complaint experience $x=\Omega_C(\mathcal{K})$, while the service agent observes the available service information $b=\Omega_S(\mathcal{K})$ and partial complaint information $z_t$, where $z_t\subseteq x$. This observation gap makes information acquisition integral to complaint resolution.

\paragraph{Progressive Customer Dynamics.}

Progressive disclosure governs how the customer's private experience becomes observable over the course of the interaction. The interaction begins with an opening complaint $u_1^C$ grounded in the service experience $x$, establishing the initial observable complaint state $z_1$ without exposing the full experience. In subsequent turns, newly disclosed information $\delta_t$ is selected through a disclosure function $\Delta$ conditioned on persona $p$, expectation $e$, the underlying experience $x$, the current observable state $z_{t-1}$, and the preceding interaction $D_{<t}$.

\begin{equation}
\begin{aligned}
\delta_t &= \Delta(x,z_{t-1};p,e,D_{<t}),\\
z_t &= z_{t-1}\cup\delta_t.
\end{aligned}
\end{equation}

\paragraph{Interaction Evolution.}

Given the current observation $o_t^S$, the service policy $\pi_S$ selects a recovery response $u_t^S$ from a decision space spanning symbolic and material strategies. Each service decision interacts with the subsequent customer response to induce a transition of the latent interaction state $h_t$. Successive state transitions define the interaction rollout, which continues until the complaint reaches an acceptable resolution or the customer disengages from the interaction.

\begin{equation}
h_{t+1}=F(h_t,u_t^S,u_{t+1}^C).
\end{equation}

\begin{table*}[!t]
    \centering
    \resizebox{15cm}{!}{
    \begin{tabular}{lccccccc}
        \toprule
        \multirow{2}{*}{\textbf{Model}} 
        & \multicolumn{4}{c}{\textbf{Justice Score}} 
        & \multirow{2}{*}{\textbf{Diseng.} }
        & \multirow{2}{*}{\textbf{Agree.} }
        & \multirow{2}{*}{\textbf{Goodwill} } \\
        \cmidrule(lr){2-5}
        & {Interactional} 
        & {Procedural} 
        & {Distributive} 
        & {Total} 
        & 
        & 
        & \\
        \midrule

        \multicolumn{8}{c}{\textbf{Airline}} \\
        \midrule
        GPT-5 & \second{$3.3050_{\pm0.8922}$} & \best{$3.4950_{\pm0.7765}$} & \second{$3.8200_{\pm0.5995}$} & \second{$3.5400_{\pm0.6433}$} & \best{0.0536} & \best{0.9464} & 91.2 \\
        GPT-4o & $2.6500_{\pm0.7212}$ & $2.6400_{\pm0.6500}$ & $3.2600_{\pm0.7845}$ & $2.8500_{\pm0.6041}$ & 0.2199 & 0.7801 & 72.2 \\
        GPT-4o mini & $2.2150_{\pm0.5100}$ & $2.1750_{\pm0.5531}$ & $2.8000_{\pm0.7957}$ & $2.3967_{\pm0.5235}$ & 0.4619 & 0.5330 & 69.0 \\
        DeepSeek-V4-Pro & \best{$3.5000_{\pm0.8507}$} & \second{$3.4050_{\pm0.8688}$} & \best{$3.9150_{\pm0.5912}$} & \best{$3.6067_{\pm0.6814}$} & \second{0.0847} & \second{0.9153} & 70.2 \\
        Gemini 2.5 Flash & $2.4350_{\pm0.6842}$ & $2.4800_{\pm0.6257}$ & $3.0450_{\pm0.8524}$ & $2.6533_{\pm0.6178}$ & 0.3920 & 0.6080 & \second{67.4} \\
        Llama 3.3 70B & $2.2150_{\pm0.4470}$ & $2.2900_{\pm0.4765}$ & $2.7600_{\pm0.8638}$ & $2.4217_{\pm0.4894}$ & 0.2938 & 0.7062 & \best{25.9} \\
        Qwen3-32B & $2.7100_{\pm0.6769}$ & $2.7700_{\pm0.7068}$ & $3.3150_{\pm0.8180}$ & $2.9317_{\pm0.6174}$ & 0.1436 & 0.8564 & 97.2 \\

        \midrule
        \multicolumn{8}{c}{\textbf{Hotel}} \\
        \midrule
        GPT-5 & \second{$3.1350_{\pm0.8605}$} & \second{$3.2800_{\pm0.7840}$} & \second{$3.8600_{\pm0.6183}$} & \second{$3.4250_{\pm0.6683}$} & \best{0.0517} & \best{0.9483} & 85.3 \\
        GPT-4o & $2.6800_{\pm0.7142}$ & $2.6300_{\pm0.6673}$ & $3.2300_{\pm0.8548}$ & $2.8467_{\pm0.6496}$ & 0.2308 & 0.7692 & 76.7 \\
        GPT-4o mini & $2.1500_{\pm0.4455}$ & $2.1700_{\pm0.4923}$ & $2.6500_{\pm0.7749}$ & $2.3233_{\pm0.4689}$ & 0.3814 & 0.6186 & 67.2 \\
        DeepSeek-V4-Pro & \best{$3.4200_{\pm0.8703}$} & \best{$3.3300_{\pm0.8274}$} & \best{$3.8900_{\pm0.6081}$} & \best{$3.5467_{\pm0.6800}$} & \second{0.0952} & \second{0.8995} & 77.2 \\
        Gemini 2.5 Flash & $2.3750_{\pm0.6134}$ & $2.4100_{\pm0.6031}$ & $2.8900_{\pm0.8670}$ & $2.5583_{\pm0.5864}$ & 0.4213 & 0.5736 & \second{58.5} \\
        Llama 3.3 70B & $2.2100_{\pm0.4971}$ & $2.3100_{\pm0.5706}$ & $2.6600_{\pm0.9157}$ & $2.3933_{\pm0.5579}$ & 0.2412 & 0.7588 & \best{19.5} \\
        Qwen3-32B & $2.8400_{\pm0.7464}$ & $2.8650_{\pm0.7207}$ & $3.3750_{\pm0.8110}$ & $3.0267_{\pm0.6379}$ & 0.1168 & 0.8832 & 90.6 \\

        \midrule
        \multicolumn{8}{c}{\textbf{Restaurant}} \\
        \midrule
        GPT-5 & \second{$3.3700_{\pm0.9150}$} & \best{$3.6150_{\pm0.7547}$} & \best{$4.1100_{\pm0.4991}$} & \best{$3.6983_{\pm0.6239}$} & \best{0.0125} & \best{0.9875} & 12.7 \\
        GPT-4o & $3.0100_{\pm0.7568}$ & $2.7800_{\pm0.6957}$ & $3.7100_{\pm0.6465}$ & $3.1667_{\pm0.5788}$ & 0.1566 & 0.8384 & 13.3 \\
        GPT-4o mini & $2.4050_{\pm0.6023}$ & $2.2350_{\pm0.5303}$ & $3.1900_{\pm0.8351}$ & $2.6100_{\pm0.5295}$ & 0.3300 & 0.6600 & \second{11.1} \\
        DeepSeek-V4-Pro & \best{$3.5300_{\pm0.8014}$} & \second{$3.3750_{\pm0.8532}$} & \second{$4.0850_{\pm0.4227}$} & \second{$3.6633_{\pm0.6061}$} & \second{0.1031} & \second{0.8969} & 12.0 \\
        Gemini 2.5 Flash & $2.7050_{\pm0.7944}$ & $2.6100_{\pm0.7816}$ & $3.5800_{\pm0.7459}$ & $2.9650_{\pm0.6695}$ & 0.3850 & 0.6150 & 12.4 \\
        Llama 3.3 70B & $2.4050_{\pm0.5500}$ & $2.4550_{\pm0.5653}$ & $3.1650_{\pm0.8841}$ & $2.6750_{\pm0.5556}$ & 0.1842 & 0.8105 & \best{4.5} \\
        Qwen3-32B & $3.1300_{\pm0.7850}$ & $2.9900_{\pm0.8623}$ & $3.5400_{\pm0.7884}$ & $3.2200_{\pm0.7013}$ & 0.1055 & 0.8894 & 20.8 \\

        \midrule
        \multicolumn{8}{c}{\textbf{Retail}} \\
        \midrule
        GPT-5 & \second{$3.5300_{\pm0.8560}$} & \best{$3.6350_{\pm0.7647}$} & \second{$4.0200_{\pm0.5668}$} & \best{$3.7283_{\pm0.6323}$} & \best{0.0274} & \best{0.9726} & 26.7 \\
        GPT-4o & $2.8350_{\pm0.7943}$ & $2.6750_{\pm0.7696}$ & $3.5500_{\pm0.7418}$ & $3.0200_{\pm0.6638}$ & 0.2172 & 0.7828 & 30.7 \\
        GPT-4o mini & $2.2500_{\pm0.5468}$ & $2.2200_{\pm0.4719}$ & $2.9100_{\pm0.8459}$ & $2.4600_{\pm0.5118}$ & 0.4821 & 0.5179 & 29.7 \\
        DeepSeek-V4-Pro & \best{$3.5900_{\pm0.9089}$} & \second{$3.3850_{\pm0.9060}$} & \best{$4.0550_{\pm0.5509}$} & \second{$3.6767_{\pm0.6985}$} & 0.1392 & 0.8608 & \second{25.7} \\
        Gemini 2.5 Flash & $2.6200_{\pm0.8540}$ & $2.5450_{\pm0.7284}$ & $3.3800_{\pm0.8714}$ & $2.8483_{\pm0.7166}$ & 0.4141 & 0.5859 & 28.4 \\
        Llama 3.3 70B & $2.4100_{\pm0.6195}$ & $2.4750_{\pm0.6335}$ & $3.1250_{\pm0.9185}$ & $2.6700_{\pm0.6144}$ & 0.1798 & 0.8202 & \best{11.0} \\
        Qwen3-32B & $3.1650_{\pm0.8727}$ & $3.1050_{\pm0.8351}$ & $3.6250_{\pm0.8765}$ & $3.2983_{\pm0.7608}$ & \second{0.0567} & \second{0.9433} & 41.9 \\

        \midrule
        \multicolumn{8}{c}{\textbf{Ride-Hailing and Delivery}} \\
        \midrule
        GPT-5 & \second{$3.2250_{\pm0.8473}$} & \second{$3.4550_{\pm0.7074}$} & \second{$3.8950_{\pm0.5794}$} & \second{$3.5250_{\pm0.6129}$} & \best{0.0340} & \best{0.9660} & 24.3 \\
        GPT-4o & $2.7300_{\pm0.7550}$ & $2.6600_{\pm0.7260}$ & $3.3300_{\pm0.8334}$ & $2.9067_{\pm0.6718}$ & 0.2769 & 0.7231 & 25.1 \\
        GPT-4o mini & $2.1900_{\pm0.4299}$ & $2.1650_{\pm0.4227}$ & $2.7800_{\pm0.8154}$ & $2.3783_{\pm0.4517}$ & 0.4772 & 0.5228 & 19.6 \\
        DeepSeek-V4-Pro & \best{$3.6750_{\pm0.8907}$} & \best{$3.5100_{\pm0.8681}$} & \best{$4.0550_{\pm0.5776}$} & \best{$3.7467_{\pm0.6988}$} & \second{0.1031} & \second{0.8969} & \second{18.1} \\
        Gemini 2.5 Flash & $2.5050_{\pm0.7298}$ & $2.4300_{\pm0.6536}$ & $3.2600_{\pm0.8342}$ & $2.7317_{\pm0.6329}$ & 0.4619 & 0.5381 & 21.1 \\
        Llama 3.3 70B & $2.3200_{\pm0.5282}$ & $2.3900_{\pm0.5741}$ & $2.9350_{\pm0.9247}$ & $2.5483_{\pm0.5787}$ & 0.2784 & 0.7216 & \best{6.0} \\
        Qwen3-32B & $2.8550_{\pm0.8169}$ & $2.9350_{\pm0.7962}$ & $3.2900_{\pm0.8774}$ & $3.0267_{\pm0.7100}$ & 0.1333 & 0.8667 & 37.8 \\
        \bottomrule
    \end{tabular}
    }
    \vspace{-4pt}
    \caption{Main evaluation results on \textsc{ServiceBench} across five service industries. Justice scores are reported as mean $\pm$ standard deviation. The best and second best results are highlighted with \best{bold orange} and \second{blue} backgrounds.}
    \label{tab:servicebench_main_results}
    \vspace{-10pt}
\end{table*}

\subsection{Evaluation}
\label{sec:evaluation}

For each rollout $\tau$, \textsc{ServiceBench} evaluates resolution quality along three complementary dimensions grounded in justice theory \cite{smith1999model}. \textbf{Interactional justice} $J^{int}(\tau)$ captures the quality of interpersonal handling across the interaction, \textbf{procedural justice} $J^{pro}(\tau)$ captures the quality of the resolution process, and \textbf{distributive justice} $J^{dis}(\tau)$ captures the appropriateness of the resulting outcome relative to the service failure and customer impact. Each dimension is scored on a 1--5 scale by the evaluation model. The resolution quality is operationalized as

\begin{equation}
Q(\tau)
=
\frac{
J^{int}(\tau)
+
J^{pro}(\tau)
+
J^{dis}(\tau)
}{3}.
\end{equation}

We additionally record disengagement $d_i\in\{0,1\}$, agreement $a_i\in\{0,1\}$, and goodwill $g_i\in\mathbb{R}_{\geq 0}$, with goodwill representing the final discretionary amount offered by the service agent. These outcomes are extracted from the completed interaction by an LLM extractor.

\section{Experiment}
\subsection{Experimental Setup}

\paragraph{Agent Models.}
We evaluate multiple LLM families, including GPT-5 \cite{singh2025openai}, GPT-4o \cite{hurst2024gpt4o}, GPT-4o mini \cite{openai2024gpt4omini}, DeepSeek-V4-Pro \cite{deepseek2026v4pro}, Gemini 2.5 Flash \cite{comanici2025gemini}, Llama 3.3 70B Instruct \cite{meta2024llama33}, and Qwen3-32B \cite{yang2025qwen3}. All models are evaluated as service agents under the same simulation protocol.

\paragraph{Judge Models.}
LLM-based judges have been widely used for automatic evaluation \cite{srinivas2025substance,zhang2025intentionesc}. We adopt GPT-5 as the primary evaluation model and examine the robustness of service agent rankings using three additional judges from other LLM families, DeepSeek-V4-Pro, Llama 3.3 70B, and Qwen3-32B. We further validate the reliability of LLM-based evaluation through comparisons with human ratings. Detailed results are provided in Appendix~\ref{app:benchmark-reliability}.

\begin{figure*}[t]
    \centering
    \includegraphics[width=0.95\textwidth]{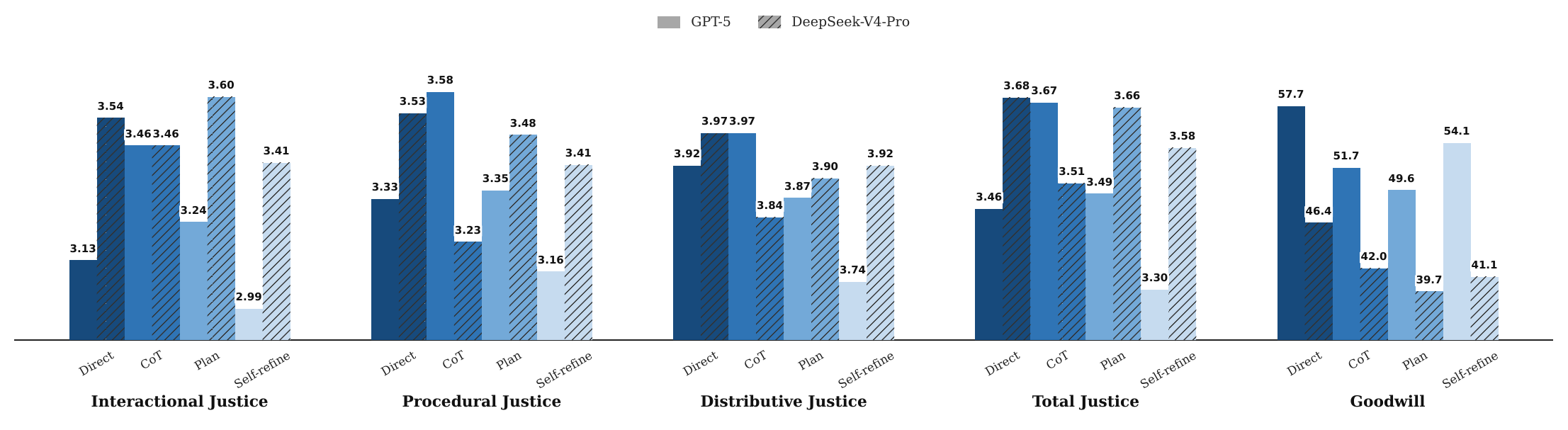}
    \caption{Performance under different inference-time reasoning strategies.}
    \label{fig:reasoning_strategies}
\end{figure*}

\begin{table*}[!t]
    \centering
    \resizebox{15cm}{!}{
    \begin{tabular}{llccccccc}
        \toprule
        \multirow{2}{*}{\textbf{Model}}
        & \multirow{2}{*}{\textbf{Guidance}}
        & \multicolumn{4}{c}{\textbf{Justice Score}}
        & \multirow{2}{*}{\textbf{Diseng.} }
        & \multirow{2}{*}{\textbf{Agree.} }
        & \multirow{2}{*}{\textbf{Goodwill} } \\
        \cmidrule(lr){3-6}
        &
        & {Interactional}
        & {Procedural}
        & {Distributive}
        & {Total}
        &
        &
        & \\
        \midrule

        \multirow{4}{*}{GPT-5}
        & None
        & 3.17 & 3.35 & 3.83 & 3.45
        & 0.025 & 0.975 & 53.80 \\

        & Case
        & 3.46 & 3.58 & 3.97 & 3.67
        & 0.013 & 0.987 & 51.74 \\

        & Procedure
        & \textbf{3.86} & \textbf{3.83} & \textbf{4.07} & \textbf{3.92}
        & \textbf{0.012} & \textbf{0.988} & 52.58 \\

        & Case + Procedure
        & 3.28 & 3.44 & 4.01 & 3.58
        & 0.042 & 0.958 & \textbf{49.62} \\

        \midrule

        \multirow{4}{*}{DeepSeek-V4-Pro}
        & None
        & 3.47 & 3.13 & 3.90 & 3.50
        & 0.111 & 0.889 & 49.40 \\

        & Case
        & 3.46 & 3.23 & 3.84 & 3.51
        & 0.082 & 0.918 & \textbf{41.97} \\

        & Procedure
        & \textbf{3.88} & \textbf{3.70} & \textbf{4.12} & \textbf{3.90}
        & \textbf{0.021} & \textbf{0.979} & 45.95 \\

        & Case + Procedure
        & 3.53 & 3.38 & 4.01 & 3.64
        & 0.092 & 0.908 & 43.95 \\

        \midrule

        \multirow{4}{*}{Llama 3.3 70B}
        & None
        & 2.20 & 2.30 & 2.72 & 2.41
        & 0.255 & 0.734 & 10.66 \\

        & Case
        & 2.44 & 2.29 & 3.04 & 2.59
        & \textbf{0.122} & \textbf{0.878} & 19.24 \\

        & Procedure
        & \textbf{2.71} & \textbf{2.65} & \textbf{3.16} & \textbf{2.84}
        & 0.140 & 0.860 & 19.80 \\

        & Case + Procedure
        & 2.26 & 2.37 & 2.84 & 2.49
        & 0.233 & 0.767 & \textbf{8.65} \\

        \midrule

        \multirow{4}{*}{Qwen3-32B}
        & None
        & 2.92 & 2.83 & 3.18 & 2.98
        & 0.091 & 0.909 & 71.47 \\

        & Case
        & 2.93 & 2.84 & 3.22 & 3.00
        & 0.103 & 0.897 & 71.41 \\

        & Procedure
        & \textbf{3.35} & \textbf{3.27} & \textbf{3.58} & \textbf{3.40}
        & \textbf{0.062} & \textbf{0.938} & 74.84 \\

        & Case + Procedure
        & 2.86 & 2.86 & 3.30 & 3.01
        & 0.072 & 0.928 & \textbf{64.99} \\

        \bottomrule
    \end{tabular}
    }

    \vspace{-4pt}
    \caption{Effect of case-based and procedural guidance on complaint handling performance.}
    \label{tab:guidance}
    \vspace{-8pt}
\end{table*}

\subsection{Main Results}

Table~\ref{tab:servicebench_main_results} reports the performance of the evaluated service agents across five service industries.

\paragraph{Higher justice performance is not necessarily associated with greater goodwill expenditure.}

GPT-5 and DeepSeek-V4-Pro achieve the strongest overall performance. DeepSeek-V4-Pro slightly outperforms GPT-5 while offering 15.2\% less goodwill. Qwen3-32B ranks ahead of GPT-4o, GPT-4o mini, and Llama 3.3 70B, but records the highest goodwill expenditure among all evaluated models, 40.2\% above DeepSeek-V4-Pro.

\paragraph{Models are more effective at delivering appropriate remedies than at interpersonal and procedural handling.}

Distributive justice consistently exceeds interactional and procedural justice across the evaluated models. This gap is evident even for GPT-5, whose distributive justice scores are 18.9\% and 12.7\% higher, and becomes more pronounced for GPT-4o mini, reaching 27.8\% and 30.7\%.

\paragraph{Resolution quality and goodwill expenditure vary substantially across service domains.}

Restaurant and retail record the highest overall justice scores, averaging 6.8\% above the remaining industries. For GPT-5, average goodwill offered in airline and hotel complaints is 4.5$\times$ that in restaurant and retail.

\begin{figure*}[t]
    \centering
    \includegraphics[width=0.95\textwidth]{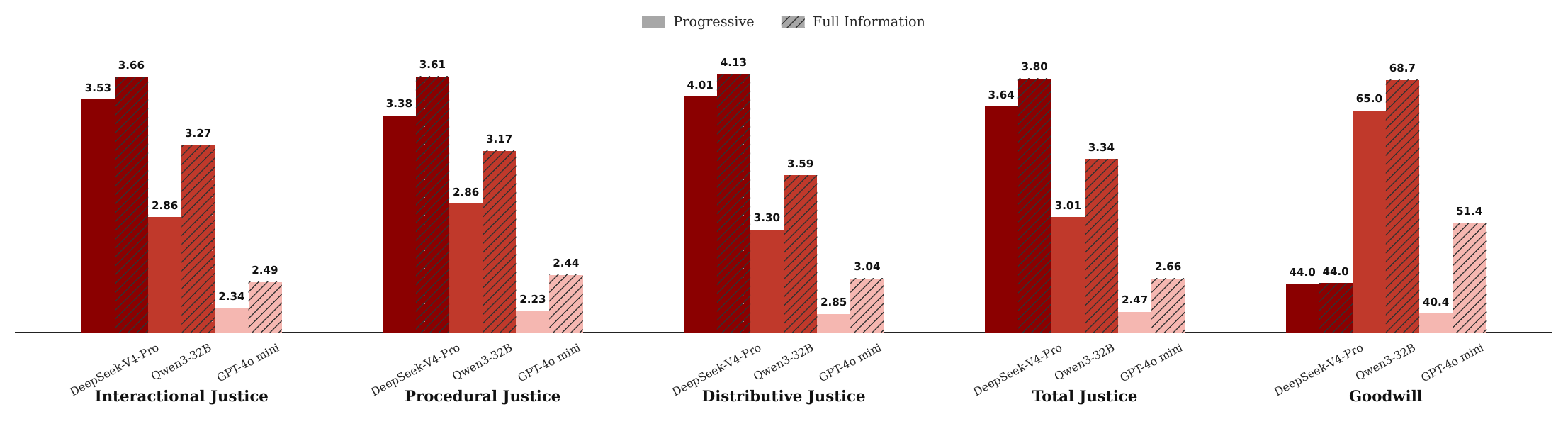}
    \caption{Performance under progressive disclosure and full information.}
    \label{fig:full_information}
\end{figure*}

\begin{table*}[t]
    \centering
    \small
    \setlength{\tabcolsep}{7pt}
    \renewcommand{\arraystretch}{1.08}

    \begin{tabular}{llccccc}
        \toprule
        \textbf{Model}
        & \textbf{Measure}
        & \textbf{Null}
        & \textbf{Turn 1}
        & \textbf{Turn 2}
        & \textbf{Turn 3+}
        & \textbf{Overall} \\
        \midrule

        \multirow{2}{*}{DeepSeek-V4-Pro}
        & First Goodwill (\%)
        & 11.80 & 30.80 & 48.20 & 9.20 & -- \\
        & Information Coverage (\%)
        & -- & 15.88 & 28.56 & 30.82 & 25.98 \\
        \midrule

        \multirow{2}{*}{Qwen3-32B}
        & First Goodwill (\%)
        & 2.00 & 92.00 & 4.80 & 1.20 & -- \\
        & Information Coverage (\%)
        & -- & 18.64 & 19.90 & 20.00 & 18.74 \\
        \midrule

        \multirow{2}{*}{GPT-4o mini}
        & First Goodwill (\%)
        & 12.00 & 62.60 & 17.00 & 8.40 & -- \\
        & Information Coverage (\%)
        & -- & 16.86 & 25.87 & 36.51 & 20.83 \\
        \bottomrule
    \end{tabular}

    \vspace{-4pt}
    \caption{Timing of the first goodwill offer and information coverage at commitment.}
    \label{tab:goodwill_coverage}
    \vspace{-8pt}
\end{table*}

\begin{table}[t]
    \centering
    \small
    \renewcommand{\arraystretch}{1.12}

    \resizebox{\columnwidth}{!}{
    \begin{tabular}{lccc}
        \toprule
        \multirow{2}{*}{\textbf{Model}}
        & \multicolumn{3}{c}{\textbf{Complaint Severity}} \\
        \cmidrule(lr){2-4}
        & \textbf{Low}
        & \textbf{Medium}
        & \textbf{High} \\
        \midrule

        GPT-4o
        & 0.19 (0.30)
        & 0.24 (0.25)
        & 0.27 (0.27) \\

        GPT-4o mini
        & 0.18 (0.10)
        & 0.21 (-0.10)
        & 0.21 (-0.34) \\

        GPT-5
        & 0.21 (0.57)
        & 0.27 (0.64)
        & 0.30 (0.53) \\

        DeepSeek-V4-Pro
        & 0.18 (0.00)
        & 0.22 (-0.04)
        & 0.29 (0.47) \\

        Gemini 2.5 Flash
        & 0.17 (-0.13)
        & 0.20 (-0.31)
        & 0.24 (-0.02) \\

        Llama 3.3 70B
        & 0.06 (-2.03)
        & 0.09 (-1.85)
        & 0.06 (-2.02) \\

        Qwen3-32B
        & 0.25 (1.20)
        & 0.32 (1.41)
        & 0.35 (1.12) \\

        \bottomrule
    \end{tabular}
    }

    \vspace{-4pt}
    \caption{Normalized goodwill across complaint severity levels. Parentheses report within-severity z-scores.}
    \label{tab:goodwill_severity}
    \vspace{-8pt}
\end{table}

\section{In-Depth Analysis}

The main results show that current service agents still struggle with important aspects of complaint handling. We therefore conduct five targeted experiments to better understand the behavioral mechanisms behind these failures and the conditions under which agent performance changes.

\subsection{Inference-Time Reasoning}
\label{sec:inference-reasoning}

Inference-time reasoning is widely used to improve agent performance, but its effect on complaint handling is less clear. We compare Chain-of-Thought (CoT) \cite{wei2022chain}, Plan-and-Solve \cite{wang2023plan}, and Self-refine \cite{madaan2023self} with Direct generation. We find that additional reasoning can improve complaint handling, but the gains depend strongly on both the model and the reasoning strategy (Figure~\ref{fig:reasoning_strategies}). GPT-5 benefits more consistently from CoT and Plan-and-Solve. Self-refine, however, consistently lowers justice scores and can also increase goodwill expenditure. More interestingly, refinement often makes an initially reasonable response worse. It can stall the dialogue or move the response away from the customer's main resolution objective.

\subsection{Case-Based and Procedural Guidance}
\label{sec:guidance}

Complaint handling often relies on organizational knowledge beyond the ongoing customer interaction. Service agents may draw on prior cases that document handling decisions or on operating procedures that provide guidance for resolution. We therefore construct representative complaint cases and SOPs based on industry examples and use them as case based and procedural guidance. Both forms of guidance improve performance, but we find that procedural guidance is consistently effective. Interestingly, combining case based and procedural guidance does not yield further improvements. Since human service agents commonly work with both, we think this drop may come from the model struggling to reconcile specific case examples with general procedural instructions, leading to less consistent decisions.

\subsection{Full-Information Oracle}

Recent work has found that LLM agents often show passive information gathering behavior when key information is missing, which can in turn degrade downstream task performance \cite{huang2025proactive}. We ask whether a similar problem appears in complaint handling. To test this, we compare our progressive disclosure setting, where the agent must gradually learn the customer's experience through the conversation, with a full-information oracle that provides the complete experience from the first turn. We find that removing this information gap improves all three justice dimensions and also leads to moderately higher goodwill expenditure (Figure~\ref{fig:full_information}). More importantly, we observe a clear change in agent behavior. With full information, agents engage more directly with the complaint and respond with greater empathy. Under progressive disclosure, they are more likely to move toward an answer before the case is sufficiently understood. As a result, the conversation can remain polite and cooperative on the surface while gathering little new information and making limited progress toward resolution.

\begin{figure*}[t]
    \centering
    \includegraphics[width=\textwidth]{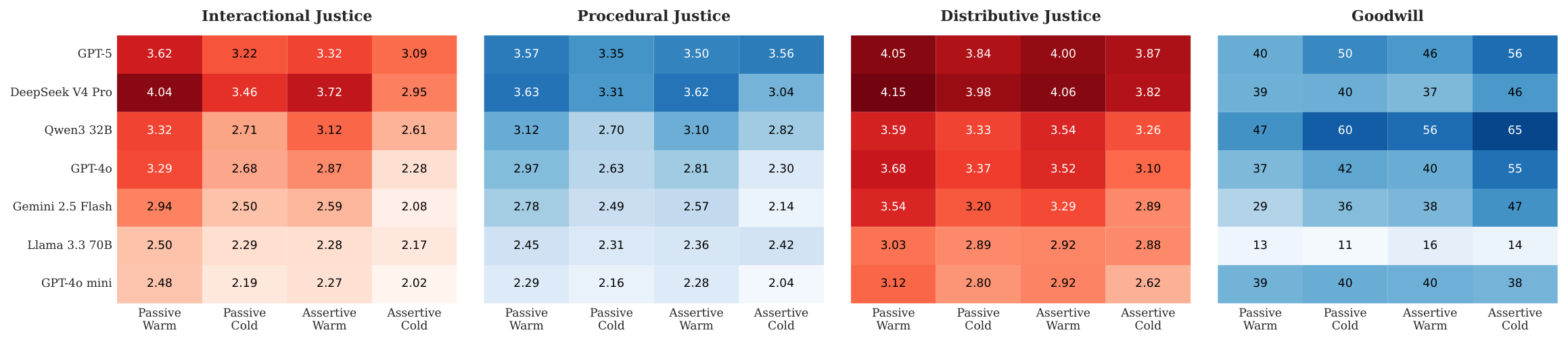}
    \caption{
    Model performance across customer communication styles.
    }
    \label{fig:communication_style_justice_heatmap}
\end{figure*}

\begin{table*}[t]
    \centering
    \small
    \renewcommand{\arraystretch}{1.10}
    \setlength{\tabcolsep}{6pt}

    \resizebox{0.98\linewidth}{!}{
    \begin{tabular}{lccccccc}
        \toprule
        \multirow{2}{*}{\textbf{Model}}
        & \multicolumn{4}{c}{\textbf{Justice Score}}
        & \multirow{2}{*}{\textbf{Diseng.} }
        & \multirow{2}{*}{\textbf{Agree.} }
        & \multirow{2}{*}{\textbf{Goodwill} } \\
        \cmidrule(lr){2-5}
        & \textbf{Interactional}
        & \textbf{Procedural}
        & \textbf{Distributive}
        & \textbf{Total}
        &
        &
        & \\
        \midrule

        GPT-5
        & 3.11
        & 3.37
        & 3.84
        & 3.44
        & 0.053
        & 0.947
        & 57.95 \\

        DeepSeek-V4-Pro
        & 3.27
        & 3.18
        & \textbf{3.95}
        & 3.47
        & 0.215
        & 0.785
        & 57.40 \\

        \midrule

        Qwen3-8B
        & 2.85
        & 2.76
        & 3.23
        & 2.95
        & 0.214
        & 0.786
        & 86.25 \\

        \quad + SFT
        & 3.39
        & 3.34
        & \textbf{3.68}
        & 3.47
        & 0.093
        & 0.907
        & 51.05 \\

        \quad + SFT + DPO
        & \textbf{3.69}
        & \textbf{3.46}
        & 3.60
        & \textbf{3.58}
        & \textbf{0.042}
        & \textbf{0.958}
        & \textbf{44.59} \\

        \bottomrule
    \end{tabular}
    }

    \caption{Effect of targeted post-training on Qwen3-8B complaint handling performance.}
    \label{tab:model_adaptation}
\end{table*}

\subsection{Goodwill Timing and Calibration}
\label{sec:goodwill}

Goodwill is an important decision in complaint handling because it can acknowledge customer inconvenience and support resolution, but inappropriate use can undermine the interaction. We therefore examine when service agents choose to offer goodwill. For each interaction, we identify the first goodwill turn and measure how much complaint information has been disclosed by that point, with calculation details provided in Appendix~\ref{app:goodwill-information}. We find that agents commit to goodwill surprisingly early, often within the first two turns when only a small portion of the customer experience has been revealed. We refer to this behavior as \textit{premature goodwill commitment}, where agents use early goodwill as an easy route to complaint closure instead of fully engaging with the underlying issue.

We then ask whether this behavior comes from a poor understanding of complaint severity. We group complaint categories into multiple severity levels and compare the goodwill offered across these groups. Most models increase goodwill as severity rises, showing that they can distinguish the relative seriousness of different service failures. This suggests that current LLMs already have an understanding of complaint scenarios. The remaining gap is therefore less about recognizing the complaint and more about turning that understanding into effective multi-turn handling behavior.

\subsection{Communication Style}
\label{sec:communication-style}

To examine how service agents respond to different customer communication patterns, we follow the Interpersonal Circumplex \cite{wiggins1991agency} and construct four styles along the agency and communion dimensions. Across these settings, we see agents struggle in opposite ways (Figure~\ref{fig:communication_style_justice_heatmap}). When customers are passive, agents often remain reactive and fail to take the initiative. This is similar to the weak mixed-initiative control reported in prior work on task oriented agents \cite{dong2025protod}, where agents have limited ability to move the conversation forward on their own. However, we also find that highly aggressive customers can push agents in the opposite direction. Agents may over accommodate user pressure and be pulled away from an appropriate handling strategy. We argue that even as models become more capable, current agents still lack robust interaction policy control and do not reliably know when to take initiative and when to resist user pressure.

\section{Model Adaptation for Complaint Handling}
\label{sec:model_adaptation}

Our analyses reveal several recurring weaknesses in complaint handling that are not simply explained by general model capability. We therefore ask whether these behaviors can be improved through targeted post-training. We fine-tune Qwen3-8B with supervised fine-tuning (SFT), followed by direct preference optimization (DPO) \cite{rafailov2023direct}. We first split the scenarios into separate training and evaluation sets. The SFT data are manually selected from existing rollouts to retain high quality examples. The DPO preference pairs are then constructed from responses generated by the SFT model and manually screened to target the behavioral limitations identified in our earlier analyses. Full data construction and training details are provided in Appendix~\ref{app:post-training}.

We find that targeted post-training leads to substantial improvements in complaint handling performance (Table~\ref{tab:model_adaptation}). After SFT and DPO, Qwen3-8B matches or surpasses substantially larger general purpose models on several evaluation outcomes despite its smaller scale. At the same time, goodwill expenditure decreases, showing that better complaint handling does not require heavier reliance on material concessions. These results suggest that the behavioral weaknesses are trainable and can be improved through task specific adaptation.

\section{Conclusion}

We introduced \textsc{ServiceBench}, a multi-turn benchmark for complaint handling, together with Asymmetric Progressive Complaint Simulation (APCS) and a justice-based evaluation framework. Our experiments show that current LLM agents have clear understanding on service scenarios, but still face clear behavioral limitations in multi-turn complaint handling. We further find that targeted post-training can substantially reduce these limitations and improve overall performance. We hope \textsc{ServiceBench} provides a useful testbed for developing more capable service agents.

\section*{Limitations}

\textsc{ServiceBench} has several limitations. First, the benchmark is primarily grounded in English complaint dialogues associated with major U.S. brands and customer profiles. This design supports consistency in scenario construction and evaluation, but it may not fully capture multilingual complaint behavior or regional variation in service norms. Expectations about apology, explanation, escalation, and remedy can differ across linguistic and cultural contexts. Extending \textsc{ServiceBench} to more languages, regions, and customer populations would provide a stronger test of cross-cultural and cross-lingual generalization.

Second, although \textsc{ServiceBench} covers five common service industries, its scope is not exhaustive. The benchmark focuses on consumer facing service contexts, whereas other domains may involve different complaint protocols. Extending the benchmark to these settings would help assess generalization beyond common consumer service environments.

Finally, \textsc{ServiceBench} is designed as a general benchmark for evaluating LLM complaint handling capability. While it supports comparisons across LLM agents, the absence of a human service agent baseline limits our ability to assess their performance relative to human practitioners. Moreover, the scenarios do not model the full policy manuals or backend systems of individual companies. Future work can incorporate human agent comparisons, company specific policies, transactional records, and tool/API use to better assess complaint handling in real world service environments.

\section*{Ethical Considerations}

The source complaint dialogues used in this study are publicly available. We do not release raw posts, user handles, URLs, or other personally identifiable information. The final benchmark cases are de-identified and manually reviewed and revised by the research team to reduce privacy risks and remove privacy sensitive or harmful content.

Our study examines complaint handling by LLM agents, a setting that involves subjective judgments about fairness, responsibility, explanation, and remedy. We acknowledge that the benchmark and evaluation metrics may reflect biases from English language complaints, major consumer service brands, and culturally specific expectations about complaint handling. We mitigate these risks through broad industry coverage, diverse complaint categories, domain informed revision, and human validation. However, biases may still persist.

While \textsc{ServiceBench} aims to support more reliable evaluation of complaint handling agents, it could be misused to justify fully automated complaint handling in consequential service settings. We strongly discourage such use without human oversight, company specific policy validation, and safeguards for affected customers.

\bibliography{custom}

\appendix

\section{Additional Experiments and Analysis}
\label{app:additional-experiments}

\subsection{Self-Preference}
\label{app:self-preference}

An important question in LLM based evaluation is whether a judge treats outputs from its own model family differently from those generated by other models. We investigate this effect by comparing each evaluator's relative scoring tendency across service agent families. Since the four judges differ in their use of the rating scale, we first standardize their scores before measuring relative preference. For dialogue $i$ evaluated by judge $j$, we define the normalized preference advantage as

\begin{equation}
d_{ij}
=
\frac{s_{ij}-\mu_j}{\sigma_j}
-
\frac{1}{3}
\sum_{k\neq j}
\frac{s_{ik}-\mu_k}{\sigma_k},
\end{equation}

\begin{table*}[t]
    \centering
    \small
    \renewcommand{\arraystretch}{1.12}

    \resizebox{0.99\linewidth}{!}{
    \begin{tabular}{lccccccc}
        \toprule
        \multicolumn{1}{c}{\textbf{Judge}}
        & \textbf{DeepSeek-V4}
        & \textbf{GPT-5}
        & \textbf{Llama 70B}
        & \textbf{Qwen 32B}
        & \textbf{Gemini 2.5}
        & \textbf{GPT-4o}
        & \textbf{GPT-4o mini} \\
        \midrule

        DeepSeek-V4-Pro
        & \textbf{0.15$^{**}$ (.009)}
        & 0.13$^{*}$ (.014)
        & -0.18$^{**}$ (.002)
        & -0.19$^{**}$ (.002)
        & 0.12$^{*}$ (.019)
        & -0.04 (.388)
        & 0.01 (.916) \\

        GPT-5
        & 0.07 (.131)
        & \textbf{0.09$^{*}$ (.024)}
        & 0.03 (.505)
        & -0.06 (.088)
        & -0.03 (.305)
        & 0.03 (.410)
        & -0.12$^{**}$ (.005) \\

        Llama 3.3 70B
        & -0.19$^{***}$ ($<$.001)
        & -0.22$^{***}$ ($<$.001)
        & \textbf{0.27$^{***}$ ($<$.001)}
        & 0.07 (.158)
        & 0.04 (.428)
        & -0.03 (.533)
        & 0.06 (.246) \\

        Qwen3-32B
        & -0.04 (.439)
        & 0.00 (.986)
        & -0.11$^{**}$ (.007)
        & \textbf{0.19$^{***}$ ($<$.001)}
        & -0.13$^{***}$ ($<$.001)
        & 0.04 (.351)
        & 0.05 (.258) \\

        \bottomrule
    \end{tabular}
    }

    \vspace{-4pt}
    \caption{
    Normalized judge preference advantage. Statistical significance is denoted by $^{*}=p<0.05$, $^{**}=p<0.01$, and $^{***}=p<0.001$.
    }
    \label{tab:judge_self_preference}
    \vspace{-6pt}
\end{table*}

\begin{table*}[t]
    \centering
    \small
    \renewcommand{\arraystretch}{1.10}

    \resizebox{0.99\linewidth}{!}{
    \begin{tabular}{lccccccc}
        \toprule
        \multirow{2}{*}{\textbf{Judge}}
        & \multicolumn{7}{c}{\textbf{Service Agent}} \\
        \cmidrule(lr){2-8}
        & \textbf{GPT-5}
        & \textbf{DeepSeek-V4-Pro}
        & \textbf{Qwen3-32B}
        & \textbf{GPT-4o}
        & \textbf{Llama 3.3 70B}
        & \textbf{Gemini 2.5 Flash}
        & \textbf{GPT-4o mini} \\
        \midrule

        DeepSeek-V4-Pro
        & \textbf{3.64}{\scriptsize $\pm$0.62}
        & 3.58{\scriptsize $\pm$0.66}
        & 3.14{\scriptsize $\pm$0.73}
        & 2.94{\scriptsize $\pm$0.65}
        & 2.56{\scriptsize $\pm$0.58}
        & 2.75{\scriptsize $\pm$0.70}
        & 2.41{\scriptsize $\pm$0.51} \\

        GPT-5
        & \textbf{3.92}{\scriptsize $\pm$0.47}
        & 3.84{\scriptsize $\pm$0.51}
        & 3.53{\scriptsize $\pm$0.69}
        & 3.30{\scriptsize $\pm$0.69}
        & 3.00{\scriptsize $\pm$0.72}
        & 2.95{\scriptsize $\pm$0.73}
        & 2.64{\scriptsize $\pm$0.71} \\

        Llama 3.3 70B
        & \textbf{4.35}{\scriptsize $\pm$0.69}
        & 4.29{\scriptsize $\pm$0.82}
        & 4.22{\scriptsize $\pm$0.84}
        & 3.78{\scriptsize $\pm$1.04}
        & 3.65{\scriptsize $\pm$1.01}
        & 3.46{\scriptsize $\pm$1.09}
        & 3.15{\scriptsize $\pm$1.09} \\

        Qwen3-32B
        & \textbf{4.48}{\scriptsize $\pm$0.66}
        & 4.36{\scriptsize $\pm$0.84}
        & 4.25{\scriptsize $\pm$0.93}
        & 3.76{\scriptsize $\pm$1.04}
        & 3.25{\scriptsize $\pm$0.94}
        & 3.22{\scriptsize $\pm$1.02}
        & 3.03{\scriptsize $\pm$1.00} \\

        \midrule

        Ensemble
        & \textbf{4.10}{\scriptsize $\pm$0.52}
        & 4.02{\scriptsize $\pm$0.62}
        & 3.79{\scriptsize $\pm$0.71}
        & 3.44{\scriptsize $\pm$0.79}
        & 3.11{\scriptsize $\pm$0.74}
        & 3.10{\scriptsize $\pm$0.83}
        & 2.81{\scriptsize $\pm$0.76} \\

        \bottomrule
    \end{tabular}
    }

    \vspace{-4pt}
    \caption{Cross-judge consistency of service-agent evaluation. Values are reported as mean $\pm$ standard deviation.}
    \label{tab:cross_judge}
    \vspace{-8pt}
\end{table*}

\begin{table}[t]
    \centering
    \small
    \renewcommand{\arraystretch}{1.12}

    \resizebox{\columnwidth}{!}{
    \begin{tabular}{lccc}
        \toprule
        \textbf{Justice Dimension}
        & $\boldsymbol{\hat{\beta}}$
        & \textbf{$t$-statistic}
        & \textbf{$p$-value} \\
        \midrule

        Interactional
        & $-0.554^{***}$
        & $-18.06$
        & $<0.001$ \\

        Procedural
        & $-0.236^{***}$
        & $-7.58$
        & $<0.001$ \\

        Distributive
        & $-0.066^{*}$
        & $-2.06$
        & $0.040$ \\

        Overall
        & $-0.285^{***}$
        & $-10.89$
        & $<0.001$ \\

        \bottomrule
    \end{tabular}
    }

    \caption{
    Regression estimates for the association between service agent response length and justice scores.
    Statistical significance is denoted by $^{*}=p<0.05$, $^{**}=p<0.01$, and $^{***}=p<0.001$.
    }
    \label{tab:verbosity_bias}
\end{table}

where $s_{ij}$ denotes judge $j$'s score for dialogue $i$, and $\mu_j$ and $\sigma_j$ denote the mean and standard deviation of judge $j$'s scores over all evaluated dialogues. The measure compares the standardized score assigned by judge $j$ with the mean standardized score from the other three evaluators. We then aggregate $d_{ij}$ by service agent model and use a two sided one sample $t$-test to determine whether the mean preference advantage differs from zero. A positive and significant advantage for outputs from the judge's own model family constitutes evidence of self preference.

Table~\ref{tab:judge_self_preference} reveals a consistent self preference effect in complaint handling evaluation. Every judge assigns its largest preference advantage to outputs from its own model family, and all four effects are statistically significant. The magnitude, however, differs substantially across evaluators. Self preference is strongest for Llama 3.3 70B and Qwen3-32B, followed by DeepSeek-V4-Pro, while GPT-5 shows a much smaller effect. The judges also exhibit positive and negative preferences toward outputs from other model families, indicating that evaluator preference extends beyond the same family effect. GPT-5 is distinctive in showing both the weakest self preference and the smallest overall deviations in preference across model families.

\subsection{Cross-Judge Consistency}
\label{app:cross-judge-consistency}

We examine whether evaluator self-preference can materially affect model comparison. We rescore the same service agent outputs using four individual LLM judges and also compare their results with an ensemble score. For each evaluator, we compute the mean justice score of every service agent and compare the resulting rankings, using Kendall's $W$ to quantify agreement. Table~\ref{tab:cross_judge} shows noticeable differences in absolute score calibration across judges. Llama 3.3 70B and Qwen3-32B tend to assign higher scores, while DeepSeek-V4-Pro and GPT-5 use lower score ranges. However, the relative ranking of service agents remains broadly consistent across individual judges and the ensemble, with Kendall's $W=0.767$ computed from scenario-wise model rankings. GPT-5 and DeepSeek-V4-Pro remain among the strongest models, and the relative positions of the other agents also change only modestly. 

\begin{table*}[t]
    \centering
    \small
    \renewcommand{\arraystretch}{1.10}
    \setlength{\tabcolsep}{5pt}

    \resizebox{0.99\linewidth}{!}{
    \begin{tabular}{llccccccc}
        \toprule
        \multirow{2}{*}{\textbf{Customer Simulator}}
        & \multirow{2}{*}{\textbf{Service Agent}}
        & \multicolumn{4}{c}{\textbf{Justice}}
        & \multirow{2}{*}{\textbf{Goodwill}}
        & \multirow{2}{*}{\textbf{Turns}}
        & \multirow{2}{*}{\textbf{Avg. Length}} \\
        \cmidrule(lr){3-6}
        &
        & \textbf{Interactional}
        & \textbf{Procedural}
        & \textbf{Distributive}
        & \textbf{Total}
        &
        &
        & \\
        \midrule

        \multirow{7}{*}{DeepSeek-V4-Pro}
        & GPT-4o
        & 2.87 & 2.65 & 3.42 & 2.98
        & 45.0 & 6.78 & 66.39 \\

        & GPT-4o mini
        & 2.34 & 2.20 & 2.93 & 2.49
        & 40.2 & 6.38 & 56.01 \\

        & GPT-5
        & 3.25 & \textbf{3.46} & 3.88 & 3.53
        & 36.0 & 9.06 & 60.17 \\

        & DeepSeek-V4-Pro
        & \textbf{3.70} & 3.18 & \textbf{4.04} & \textbf{3.64}
        & 44.1 & 5.69 & 58.70 \\

        & Gemini 2.5
        & 3.22 & 2.85 & 3.63 & 3.23
        & 33.9 & 6.51 & 56.56 \\

        & Llama 70B
        & 2.32 & 2.22 & 2.97 & 2.50
        & \textbf{13.0} & 9.44 & 41.33 \\

        & Qwen3-32B
        & 2.72 & 2.49 & 3.13 & 2.78
        & 57.1 & 5.71 & 53.50 \\

        \midrule

        \multirow{7}{*}{GPT-5}
        & GPT-4o
        & 2.72 & 2.62 & 3.38 & 2.91
        & 49.1 & 7.75 & 68.51 \\

        & GPT-4o mini
        & 2.23 & 2.15 & 2.80 & 2.39
        & 42.9 & 7.98 & 56.54 \\

        & GPT-5
        & 3.28 & \textbf{3.47} & \textbf{4.02} & 3.59
        & 52.9 & 10.77 & 61.56 \\

        & DeepSeek-V4-Pro
        & \textbf{3.47} & 3.42 & \textbf{4.02} & \textbf{3.64}
        & 42.9 & 6.54 & 63.57 \\

        & Gemini 2.5
        & 2.52 & 2.49 & 3.19 & 2.73
        & 36.8 & 8.15 & 51.61 \\

        & Llama 70B
        & 2.36 & 2.37 & 2.91 & 2.55
        & \textbf{13.0} & 10.24 & 41.96 \\

        & Qwen3-32B
        & 2.92 & 2.94 & 3.41 & 3.09
        & 57.5 & 6.73 & 52.83 \\

        \bottomrule
    \end{tabular}
    }

    \caption{Service agent performance under different customer simulation policies.}
    \label{tab:interaction-variability}
\end{table*}

\begin{table*}[t]
    \centering
    \small
    \setlength{\tabcolsep}{8pt}
    \renewcommand{\arraystretch}{1.08}

    \begin{tabularx}{\textwidth}{X c X c}
        \toprule
        \multicolumn{2}{c}{\textbf{Top 10 Categories}}
        & \multicolumn{2}{c}{\textbf{Bottom 10 Categories}} \\
        \cmidrule(lr){1-2}
        \cmidrule(lr){3-4}

        \textbf{Category}
        & \textbf{Justice}
        & \textbf{Category}
        & \textbf{Justice} \\
        \midrule

        Product Hazard Concerns
        & 3.80
        & Baggage Follow-up Failures
        & 2.50 \\

        Delivery Handling Failures
        & 3.52
        & Room Assignment Failures
        & 2.58 \\

        Product Mismatch
        & 3.50
        & Pest and Infestation Problems
        & 2.60 \\

        Replacement Execution Failures
        & 3.48
        & Room Environment Problems
        & 2.65 \\

        Unsafe Consumables
        & 3.45
        & Specialist Service Delays
        & 2.67 \\

        Inflight Connectivity Failures
        & 3.32
        & Account Restriction Failures
        & 2.67 \\

        Charge Discrepancies
        & 3.32
        & Live Support Access Failures
        & 2.67 \\

        Emergency Escalation Failures
        & 3.31
        & Lounge Access Failures
        & 2.68 \\

        Account Resolution Failures
        & 3.31
        & Noise Disruption
        & 2.68 \\

        Registry Fulfillment Failures
        & 3.30
        & Fare Integrity Issues
        & 2.69 \\

        \bottomrule
    \end{tabularx}

    \vspace{-4pt}
    \caption{Highest and lowest performing complaint categories by total justice.}
    \label{tab:category_extremes}
    \vspace{-8pt}
\end{table*}

\begin{figure}[t]
    \centering
    \includegraphics[width=\columnwidth]
    {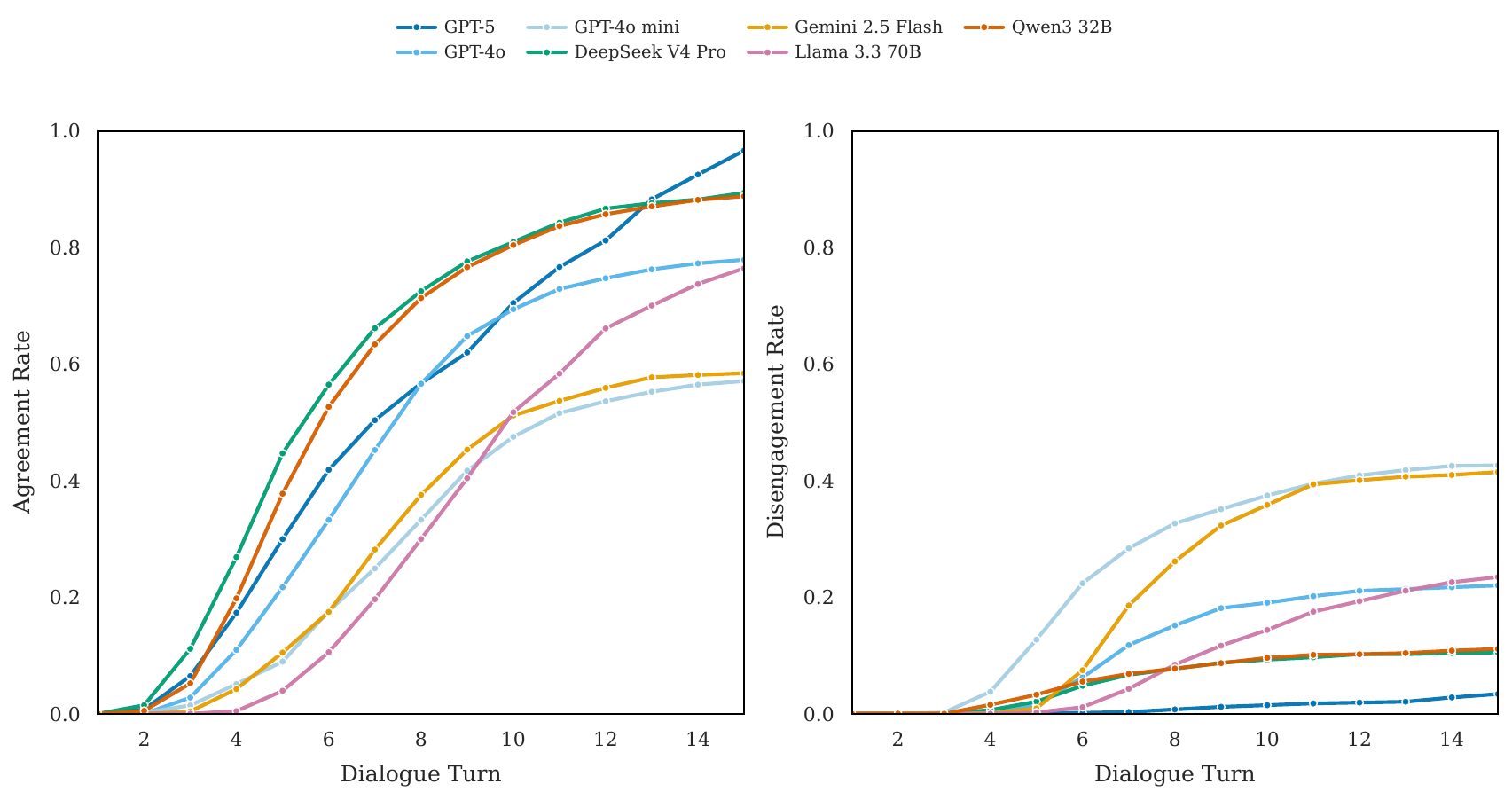}
    \caption{Cumulative resolution outcomes over dialogue turns.}
    \label{fig:cumulative_outcomes_by_turn}
\end{figure}

\subsection{Verbosity Bias}
\label{app:verbosity-bias}

Response length can act as a superficial signal of effort or completeness in LLM evaluation, raising the possibility that more verbose service agents may receive higher scores independent of their actual complaint handling quality. We examine the relationship between response length and justice evaluation using the GPT-5 judge. For each interaction, verbosity is measured by the total number of words generated by the service agent. We regress each justice score on the logarithm of response length while controlling for service agent and scenario fixed effects:

\begin{equation}
\begin{aligned}
Justice_{im} ={}& \alpha + \beta \log(W_{im}) \\
                 &+ \gamma_m + \delta_i + \epsilon_{im},
\end{aligned}
\end{equation}

where $W_{im}$ denotes the total number of words generated by model $m$ in scenario $i$. The model fixed effects $\gamma_m$ account for systematic differences in response style across service agents, while the scenario fixed effects $\delta_i$ control for variation in complaint context and difficulty. Standard errors are clustered at the scenario level. The coefficient $\beta$ therefore captures the association between response length and justice scores after accounting for these model and scenario differences.

Table~\ref{tab:verbosity_bias} shows that longer responses do not receive a scoring advantage. Instead, response length is negatively associated with all three justice dimensions and total justice. The effect is particularly pronounced for interactional justice, suggesting that excessive verbosity can signal weaker communication quality, such as inefficient information delivery or failure to advance the dialogue effectively. The smaller effect on distributive justice is also intuitive, as the substantive adequacy of a remedy depends less directly on how much the agent says. These results suggest that the evaluator does not equate greater response volume with better complaint handling. Instead, its scores appear sensitive to whether communication is efficient and effective within the service interaction.

\begin{table*}[t]
    \centering
    \small
    \renewcommand{\arraystretch}{1.10}
    \setlength{\tabcolsep}{7pt}

    \resizebox{0.98\linewidth}{!}{
    \begin{tabular}{lccccc}
        \toprule
        \textbf{Model}
        & \textbf{Airline}
        & \textbf{Hotel}
        & \textbf{Restaurant}
        & \textbf{Retail}
        & \textbf{Ride-Hailing \& Delivery} \\
        \midrule

        GPT-4o
        & 0.361 (0.076)
        & 0.383 (0.371)
        & 0.066 (0.186)
        & 0.154 (0.330)
        & 0.125 (0.354) \\

        GPT-4o mini
        & 0.345 (-0.065)
        & 0.336 (-0.027)
        & 0.055 (-0.278)
        & 0.149 (0.218)
        & 0.098 (-0.224) \\

        GPT-5
        & 0.456 (0.908)
        & 0.426 (0.731)
        & 0.064 (0.061)
        & 0.133 (-0.116)
        & 0.122 (0.274) \\

        DeepSeek-V4-Pro
        & 0.351 (-0.009)
        & 0.386 (0.393)
        & 0.060 (-0.080)
        & 0.129 (-0.221)
        & 0.090 (-0.381) \\

        Gemini 2.5 Flash
        & 0.337 (-0.134)
        & 0.292 (-0.393)
        & 0.062 (-0.006)
        & 0.142 (0.075)
        & 0.106 (-0.059) \\

        Llama 3.3 70B
        & 0.130 (-1.945)
        & 0.098 (-2.027)
        & 0.023 (-1.655)
        & 0.055 (-1.836)
        & 0.030 (-1.655) \\

        Qwen3-32B
        & 0.486 (1.168)
        & 0.453 (0.953)
        & 0.104 (1.772)
        & 0.209 (1.551)
        & 0.189 (1.692) \\

        \bottomrule
    \end{tabular}
    }

    \vspace{-4pt}
    \caption{
    Normalized goodwill expenditure across service industries.
    Values are normalized by the USD~200 discretionary limit; parentheses report within-industry standardized scores.
    }
    \label{tab:goodwill_industry}
    \vspace{-8pt}
\end{table*}

\begin{table}[t]
    \centering
    \small
    \setlength{\tabcolsep}{14pt}
    \renewcommand{\arraystretch}{0.90}
    \begin{tabular}{lcc}
        \toprule
        \textbf{Model}
        & \textbf{Turns}
        & \textbf{Length} \\
        \midrule
        GPT-5              & 10.32 & 61.74 \\
        GPT-4o             & 7.54  & 68.40 \\
        GPT-4o mini        & 7.77  & 56.66 \\
        DeepSeek-V4-Pro    & 6.69  & 63.61 \\
        Gemini 2.5 Flash   & 8.03  & 51.37 \\
        Llama 3.3 70B      & 10.13 & 42.18 \\
        Qwen3-32B          & 6.79  & 53.76 \\
        \bottomrule
    \end{tabular}

    \vspace{-4pt}
    \caption{Conversation statistics across evaluated service agents.}
    \label{tab:dialogue_efficiency}
    \vspace{-6pt}
\end{table}

\subsection{Stochastic Interaction Variability}
\label{app:interaction-variability}

The behavior of a simulated customer can alter the course of the interaction through which a service agent encounters and resolves the same underlying complaint. We examine how service agents respond to this interaction variability by replacing the GPT-5 customer simulator with DeepSeek-V4-Pro, thereby introducing a different customer response policy. All seven service agents are evaluated under both customer simulators, allowing us to compare changes in interaction characteristics and outcomes under different customer behaviors. Table~\ref{tab:interaction-variability} shows that service agent behavior remains broadly consistent across different customer simulators. Stronger agents, including DeepSeek-V4-Pro and GPT-5, are less sensitive to changes in customer behavior and maintain similar justice performance across different rollouts. Goodwill varies more noticeably, suggesting that compensation decisions depend more strongly on the specific course of the interaction.

\subsection{Resolution Dynamics}
\label{resolution-dynamics}

Complaint resolution unfolds over multiple turns, and the timing of agreement and disengagement offers insight into how service agents manage the interaction. We therefore examine the cumulative patterns of these outcomes across dialogue turns. As shown in Figure~\ref{fig:cumulative_outcomes_by_turn}, differences between models emerge early and become more pronounced as conversations progress. Weaker agents tend to encounter customer disengagement earlier, often losing the opportunity to resolve the complaint before meaningful progress is made. Yet keeping the customer engaged does not necessarily lead to agreement. An agent may sustain a conversation over many turns without adequately addressing the customer's underlying concerns or moving toward a satisfactory resolution. This distinction highlights two complementary capabilities, \textit{engagement maintenance} and \textit{resolution advancement}. Engagement maintenance reflects an agent's ability to sustain customer participation by responding to concerns, managing frustration, and preventing the conversation from breaking down prematurely. Resolution advancement goes a step further, requiring the agent to identify unresolved issues, gather relevant information, and adapt its handling strategy to make meaningful progress toward an acceptable outcome. While the former preserves the opportunity for resolution, the latter determines how effectively the agent uses that opportunity. Effective complaint handling therefore requires agents to both sustain customer engagement and actively guide the interaction toward resolution.

\begin{table*}[t]
    \centering
    \small
    \renewcommand{\arraystretch}{1.10}
    \setlength{\tabcolsep}{5.5pt}

    \resizebox{0.98\linewidth}{!}{
    \begin{tabular}{llcccccc}
        \toprule
        \multirow{2}{*}{\textbf{Model}}
        & \multirow{2}{*}{\textbf{Run}}
        & \multicolumn{4}{c}{\textbf{Justice}}
        & \multirow{2}{*}{\textbf{Goodwill}}
        & \multirow{2}{*}{\textbf{Turns}} \\
        \cmidrule(lr){3-6}
        &
        & \textbf{Interactional}
        & \textbf{Procedural}
        & \textbf{Distributive}
        & \textbf{Total}
        &
        & \\
        \midrule

        \multirow{6}{*}{DeepSeek-V4-Pro}
        & 1 & 3.47 & 3.42 & 4.02 & 3.64 & 42.9 & 6.54 \\
        & 2 & 3.43 & 3.37 & 3.98 & 3.59 & 49.2 & 6.64 \\
        & 3 & 3.52 & 3.28 & 3.98 & 3.59 & 40.9 & 6.44 \\
        & 4 & 3.45 & 3.28 & 3.99 & 3.57 & 43.3 & 6.62 \\
        & 5 & 3.38 & 3.25 & 3.83 & 3.49 & 42.2 & 6.79 \\
        & \textbf{Mean $\pm$ SD}
        & 3.45 $\pm$ 0.05
        & 3.32 $\pm$ 0.07
        & 3.96 $\pm$ 0.07
        & 3.58 $\pm$ 0.06
        & 43.7 $\pm$ 3.2
        & 6.61 $\pm$ 0.13 \\
        \midrule

        \multirow{6}{*}{GPT-5}
        & 1 & 3.28 & 3.47 & 4.02 & 3.59 & 52.9 & 10.77 \\
        & 2 & 3.27 & 3.37 & 3.86 & 3.50 & 50.9 & 10.06 \\
        & 3 & 3.26 & 3.40 & 3.85 & 3.50 & 54.7 & 9.72 \\
        & 4 & 3.28 & 3.44 & 3.88 & 3.53 & 49.4 & 10.20 \\
        & 5 & 3.22 & 3.42 & 3.86 & 3.50 & 53.1 & 10.14 \\
        & \textbf{Mean $\pm$ SD}
        & 3.26 $\pm$ 0.03
        & 3.42 $\pm$ 0.04
        & 3.89 $\pm$ 0.07
        & 3.53 $\pm$ 0.04
        & 52.2 $\pm$ 2.1
        & 10.18 $\pm$ 0.38 \\
        \midrule

        \multirow{6}{*}{Qwen3-32B}
        & 1 & 2.92 & 2.94 & 3.41 & 3.09 & 57.5 & 6.73 \\
        & 2 & 2.92 & 2.81 & 3.34 & 3.02 & 65.6 & 6.76 \\
        & 3 & 2.81 & 2.78 & 3.22 & 2.94 & 70.9 & 6.74 \\
        & 4 & 2.82 & 2.87 & 3.29 & 2.99 & 65.0 & 6.89 \\
        & 5 & 2.83 & 2.79 & 3.41 & 3.01 & 63.9 & 6.45 \\
        & \textbf{Mean $\pm$ SD}
        & 2.86 $\pm$ 0.06
        & 2.84 $\pm$ 0.07
        & 3.33 $\pm$ 0.08
        & 3.01 $\pm$ 0.06
        & 64.6 $\pm$ 4.8
        & 6.71 $\pm$ 0.16 \\
        \bottomrule
    \end{tabular}
    }

    \caption{Run-to-run consistency across five independent simulations.}
    \label{tab:run_consistency}
\end{table*}

\begin{table}[t]
    \centering
    \resizebox{\columnwidth}{!}{
    \begin{tabular}{lccc}
        \toprule
        \textbf{Annotator Pair}
        & \textbf{Interactional}
        & \textbf{Procedural}
        & \textbf{Distributive} \\
        \midrule
        Judge 1 -- Judge 2 & 0.870 & 0.903 & 0.766 \\
        Judge 1 -- Judge 3 & 0.779 & 0.810 & 0.758 \\
        Judge 2 -- Judge 3 & 0.774 & 0.763 & 0.791 \\
        \midrule
        Mean pairwise & 0.808 & 0.825 & 0.772 \\
        \bottomrule
    \end{tabular}
    }
    \vspace{-4pt}
    \caption{Human annotator agreement measured by pairwise Spearman correlation.}
    \label{tab:human_agreement}
    \vspace{-8pt}
\end{table}

\begin{figure}[t]
    \centering
    \includegraphics[width=\columnwidth]{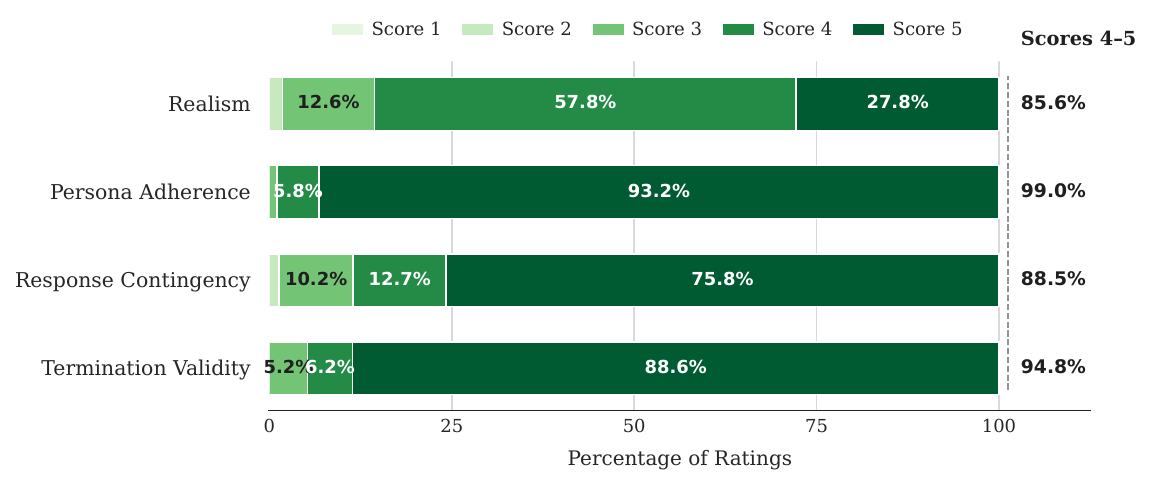}
    \caption{Human evaluation of customer simulator realism.}
    \label{fig:customer-simulator-quality}
\end{figure}

\subsection{Complaint Complexity and Agent Performance}
\label{app:category-performance}

We further examine performance across complaint categories to understand which types of service failures pose greater challenges for LLM agents. Table~\ref{tab:category_extremes} presents the ten highest and lowest scoring categories based on average total justice. Higher scoring categories typically involve problems with relatively clear resolution paths, such as product replacement, delivery correction, or charge disputes, where agents can readily identify an appropriate course of action. In contrast, lower scoring categories often involve unresolved service conditions, repeated follow-ups, or situations requiring further investigation and coordination. In these cases, the appropriate resolution is less apparent from the initial information and must be established through continued interaction. Agents need to recognize what remains unknown, proactively gather relevant information, and adjust their handling strategies as the situation develops. The differences across categories therefore reveal a broader limitation in how current agents handle complex complaint resolution. They perform relatively well when the problem is clearly defined and an appropriate remedy can be identified, but struggle when resolution requires sustained inquiry, reasoning, and coordination across multiple turns. This finding is consistent with our earlier analyses, suggesting that a key weakness lies in agents' limited ability to actively explore incomplete complaint information and translate that understanding into a coherent resolution strategy.

\begin{table}[t]
    \centering
    \small

    \resizebox{\columnwidth}{!}{
    \begin{tabular}{lcccccc}
        \toprule
        \multirow{2}{*}{\textbf{Judge Model}}
        & \multicolumn{2}{c}{\textbf{Interactional}}
        & \multicolumn{2}{c}{\textbf{Procedural}}
        & \multicolumn{2}{c}{\textbf{Distributive}} \\
        \cmidrule(lr){2-3}
        \cmidrule(lr){4-5}
        \cmidrule(lr){6-7}

        & \textbf{Spearman}
        & \textbf{MAE}
        & \textbf{Spearman}
        & \textbf{MAE}
        & \textbf{Spearman}
        & \textbf{MAE} \\
        \midrule

        GPT-5
        & 0.831 & 0.348
        & 0.813 & 0.420
        & 0.789 & 0.367 \\

        GPT-4o
        & 0.866 & 0.343
        & 0.778 & 0.468
        & 0.766 & 0.477 \\

        GPT-4o mini
        & 0.839 & 0.297
        & 0.787 & 0.378
        & 0.735 & 0.632 \\

        Llama 3.3 70B
        & 0.793 & 0.700
        & 0.778 & 0.755
        & 0.761 & 0.585 \\

        Qwen3-32B
        & 0.795 & 0.718
        & 0.764 & 0.728
        & 0.738 & 0.553 \\

        \bottomrule
    \end{tabular}
    }

    \vspace{-4pt}
    \caption{Validation of candidate LLM judges against human annotations.}
    \label{tab:judge_validation}
    \vspace{-8pt}
\end{table}

\subsection{Goodwill Timing and Information Coverage}
\label{app:goodwill-information}

We examine whether service agents offer goodwill before sufficiently understanding the complaint. For each dialogue, we identify the turn $t_i^g$ at which goodwill is first offered and measure how much complaint relevant information has been disclosed by that point. For each scenario $i$, the complete customer experience is decomposed into a set of complaint relevant information units $\mathcal{I}_i$. We then extract the dialogue preceding $t_i^g$ and identify the disclosed units $\mathcal{I}_i^g$ by matching the customer information expressed in the conversation to $\mathcal{I}_i$. Information coverage is computed as

\begin{equation}
C_i^{g}
=
\frac{|\mathcal{I}_i^{g}|}{|\mathcal{I}_i|}.
\end{equation}

The resulting ratio captures the proportion of the complete complaint experience available when the agent first commits to goodwill. We use the first goodwill turn and corresponding information coverage to compare commitment timing across service agents.

\subsection{Industry Differences in Goodwill Expenditure}
\label{app:goodwill-industry}

We examine how goodwill expenditure varies across service industries and how differences in service context may shape compensation decisions. As shown in Table~\ref{tab:goodwill_industry}, airline and hotel complaints generally receive higher goodwill amounts than those in restaurants, retail, and ride-hailing and delivery. This pattern may reflect the higher service value and greater customer commitments typically associated with airline travel and hotel stays. Service failures in these industries can disrupt broader travel or accommodation plans and impose substantial financial and experiential costs on customers. In comparison, complaints in restaurants, retail, and everyday transportation services often arise from more routine transactions with lower monetary values and less extensive consequences. These differences may explain why models tend to offer greater goodwill in airline and hotel settings.

\subsection{Conversation Statistics}
\label{app:conversation-statistics}

We examine differences in conversation length and response verbosity across service agents. Table~\ref{tab:dialogue_efficiency} reports the average number of dialogue turns and the average number of words generated by each agent per turn. GPT-5 and Llama 3.3 70B tend to sustain longer conversations, while DeepSeek-V4-Pro and Qwen3-32B typically complete interactions in fewer turns. However, conversation length and response verbosity do not follow the same pattern. GPT-4o generates the longest responses on average despite having relatively short conversations, whereas Llama 3.3 70B produces the shortest responses while engaging in some of the longest interactions. This contrast suggests that the amount of information conveyed in individual responses and the duration of an interaction reflect different aspects of conversational behavior. Neither longer responses nor more dialogue turns necessarily indicate greater progress toward complaint resolution.

\subsection{Case Studies}
\label{app:case-studies}

We present four representative case studies in Figures~\ref{fig:case-study-deepseek-accessibility}, \ref{fig:case-study-deepseek-maintenance}, \ref{fig:case-study-gpt5-reservation-change}, and \ref{fig:case-study-qwen-compensation}. These examples illustrate recurring failure patterns observed in \textsc{ServiceBench}. Current agents often generate reasonable responses at individual turns but fail to maintain a coherent resolution strategy over the full interaction. Common behaviors include premature closure around compensation, unnecessary information seeking, insufficient adaptation to unresolved customer objectives, and unstable concession decisions. These patterns highlight a broader limitation in managing the full interaction, where locally appropriate actions do not always accumulate into effective complaint resolution.

\begin{table*}[t]
\centering
\small
\setlength{\tabcolsep}{8pt}
\renewcommand{\arraystretch}{1.12}

\begin{tabularx}{\textwidth}{Y Y Y}
\toprule
\textbf{Industry} & \textbf{Selected Customer Care Accounts} & \textbf{\# Raw Source Dialogues} \\
\midrule
Airlines & AmericanAir, Delta, JetBlue, SouthwestAir & 78,864 \\
Hotels & HiltonHelp, IHGService & 2,460 \\
Restaurants & McDonalds, TacoBellTeam, KFC\_UKI\_Help & 13,247 \\
Retail & AskTarget, Walmart & 13,541 \\
Ride-Hailing and Delivery & Uber\_Support, AskLyft & 52,029 \\
\bottomrule
\end{tabularx}

\caption{Selected customer care accounts and raw dialogue counts used for customer journey map construction.}
\label{tab:selected_brands}
\end{table*}

\begin{table}[t]
    \centering
    \small
    \renewcommand{\arraystretch}{1.08}
    \begin{tabular*}{\columnwidth}{@{\extracolsep{\fill}}lc}
        \toprule
        \textbf{Benchmark Scale} & \textbf{Count} \\
        \midrule
        Scenarios & 2,500 \\
        Service industries & 5 \\
        Customer journey stages & 37 \\
        Complaint categories & 184 \\
        \midrule
        \textbf{Industry Coverage} & \textbf{Stage / Category} \\
        \midrule
        Airline & 8 / 44 \\
        Hotel & 8 / 28 \\
        Restaurant & 6 / 24 \\
        Retail & 8 / 46 \\
        Ride-hailing and delivery & 7 / 42 \\
        \midrule
        \textbf{Experience Background} & \textbf{Word Count} \\
        \midrule
        Average length & 291.29 \\
        Minimum length & 187 \\
        Maximum length & 519 \\
        \bottomrule
    \end{tabular*}
    \caption{
    Summary statistics of \textsc{ServiceBench}. Detailed complaint categories are provided in Appendix~\ref{app:customer-journey-maps}.
    }
    \label{tab:benchmark_statistics}
\end{table}

\section{Benchmark Reliability}
\label{app:benchmark-reliability}

\subsection{Customer Simulator Quality}
\label{app:customer-simulator-quality}

We assess the realism and behavioral fidelity of the customer simulator through human evaluation. Annotators with at least two years of service industry experience rate 200 sampled dialogues on overall realism, persona adherence, response contingency, and termination validity using a five-point scale. As shown in Figure~\ref{fig:customer-simulator-quality}, the simulator achieves an average realism score of 4.12, with 85.6\% of ratings at 4 or 5. Behavioral fidelity is similarly strong, with 99.0\% of persona adherence ratings, 88.5\% of response contingency ratings, and 94.8\% of termination validity ratings receiving scores of 4 or 5. These results show that the simulated customers are both contextually plausible and behaviorally consistent with the intended interaction dynamics.

\subsection{Human Agreement and Judge Selection}
\label{app:judge-selection}

To establish a human reference for the justice evaluation, three annotators independently score 200 sampled dialogues on interactional, procedural, and distributive justice using the same five-point rubric. Table~\ref{tab:human_agreement} shows strong pairwise agreement across all three dimensions, with mean Spearman correlations ranging from 0.772 to 0.825. The results demonstrate that the three justice dimensions can be applied consistently across human evaluators.

We next compare five candidate LLM judges against the aggregated human ratings. As shown in Table~\ref{tab:judge_validation}, we assess each judge using Spearman correlation for agreement in score ordering and mean absolute error (MAE) for calibration on the five-point scale. Performance varies across models and justice dimensions, but GPT-5 combines consistently strong correlations with comparatively low calibration error across all three dimensions. We therefore select GPT-5 as the primary evaluator for the benchmark.

\subsection{Run-to-Run Consistency}
\label{app:run-consistency}

We assess the consistency of benchmark outcomes under repeated stochastic simulation. A stratified sample of 100 scenarios is drawn evenly across the five service industries, and each scenario is independently simulated five times with three representative service agent models. Table~\ref{tab:run_consistency} shows only limited variation in justice scores and dialogue characteristics across repeated runs. Model performance remains consistent across simulations, supporting the reliability of the aggregate benchmark results.

\section{Additional Details On Benchmark Construction}
\label{app:benchmark-construction}

\subsection{Benchmark Statistics}
\label{app:benchmark-statistics}

\textsc{ServiceBench} contains 2,500 complaint scenarios across five service industries, organized into 37 customer journey stages and 184 complaint categories. The benchmark is constructed from real complaint dialogues through an industry specific customer journey guided pipeline that combines data driven topic discovery with expert guided coding. An overview of the benchmark statistics is provided in Table~\ref{tab:benchmark_statistics}.

\subsection{Industry and Brand Selection}
\label{Data Source and Brand Selection}

We select five service industries to capture diverse complaint handling contexts involving multiple customer journey stages, varied service failures, and recovery processes that require both interpersonal responses and discretionary material remedies. The selected industries span airlines, hotels, restaurants, retail, and ride hailing and delivery, covering a broad range of service processes, customer touchpoints, and resolution requirements. Within each industry, we select representative brands with comparable service processes and customer support models, allowing complaints from multiple brands to be organized under a unified industry level customer journey map. Although the raw source corpus is naturally imbalanced across industries, it is used only for complaint mining and taxonomy construction. The final benchmark is constructed through balanced sampling across industries and complaint categories. Table~\ref{tab:selected_brands} reports the selected customer care accounts and the number of raw complaint dialogues collected for each industry.

\begin{table*}[t]
\centering
\small
\setlength{\tabcolsep}{7pt}
\renewcommand{\arraystretch}{1.15}
\begin{tabularx}{\textwidth}{
    @{}
    >{\raggedright\arraybackslash}p{0.23\textwidth}
    >{\raggedright\arraybackslash}p{0.19\textwidth}
    >{\raggedright\arraybackslash}X
    @{}
}
\toprule
\textbf{Component} & \textbf{Provenance} & \textbf{Description} \\
\midrule
Source complaint dialogues
& Collected
& Public customer complaint dialogues collected from X, providing the original facts of service failures. \\

\addlinespace[2pt]
Customer journey maps
& Data driven and manually constructed
& Service stages and complaint categories developed through topic discovery and expert guided coding. \\

\addlinespace[2pt]
Service event sequences
& Reconstructed
& Plausible sequences of events leading to each complaint, reconstructed from source dialogues while preserving the original facts. \\

\addlinespace[2pt]
Customer experience background and resolution expectation
& Generated with manual refinement
& Detailed customer experiences and resolution expectations developed from the event sequences and refined for realism and consistency. \\

\addlinespace[2pt]
Company case record
& Generated with manual refinement
& Available company information inferred from the customer experience using industry specific schemas and manually refined for realism and consistency. \\
\bottomrule
\end{tabularx}
\caption{Provenance of the main components used in ServiceBench scenario construction.}
\label{tab:scenario-provenance}
\end{table*}

\subsection{Dialogue Preprocessing}
\label{Dialogue Preprocessing}

Raw social media dialogues contain substantial noise, including duplicated exchanges, incomplete messages, irrelevant artifacts, and negative expressions without concrete complaint information. We therefore apply a four-step preprocessing pipeline to retain dialogues suitable for complaint taxonomy construction.

\begin{itemize}

    \item \textbf{Text normalization.}
    We standardize raw text by decoding HTML entities, applying NFKC normalization, replacing URLs with \texttt{<URL>}, converting hashtags to plain text, and normalizing mentions. Numeric handles are mapped to \texttt{<USER>}, selected brand handles to \texttt{<BRAND>}, and other handles to an ``at'' form. We also remove empty parentheses, normalize ellipses and punctuation spacing, and collapse repeated whitespace.

    \item \textbf{Length and duplicate filtering.}
    We remove dialogues with five or fewer tokens and filter duplicate or near-duplicate dialogues. Near duplicates are detected using \texttt{NV-Embed-v2} \cite{lee2025nvembed} embeddings, with pairs above 0.95 cosine similarity removed.

    \item \textbf{Sentiment filtering.}
    We retain complaint candidates with negative sentiment using \texttt{twitter-roberta-base-sentiment-latest} \cite{camacho2022tweetnlp}, keeping dialogues with \(p_{\text{neg}} \geq 0.69\).

    \item \textbf{Complaint information filtering.}
    We use GPT-4o mini to identify dialogues containing concrete service failure information. A dialogue is retained only when it satisfies domain relevance, issue specificity, and content interpretability. Domain relevance requires the dialogue to concern the target service context. Issue specificity requires at least one identifiable service failure or complaint issue instead of emotion alone. Content interpretability requires the text to be understandable and sufficiently informative to infer what went wrong. Dialogues containing only vague dissatisfaction, profanity, sarcasm, or emotional reaction without an identifiable service issue are removed.

\end{itemize}

\subsection{Customer Journey Map Construction}
\label{Customer Journey Map Construction}

We construct customer journey maps through an iterative process that combines topic discovery, subtopic exploration, and expert-guided coding. BERTopic and K-means are used to surface recurring complaint patterns in the cleaned dialogue corpus. Based on these issue patterns, we develop an interpretable taxonomy that organizes service failures by industry-specific journey stages. The resulting customer journey maps provide the category structure for balanced scenario generation.

\begin{itemize}

    \item \textbf{BERTopic-based topic exploration.}
    We first apply BERTopic \cite{grootendorst2022bertopic} to identify broad recurring complaint themes from the cleaned dialogues. BERTopic combines semantic representation with density-based topic discovery and does not require the number of topics to be specified in advance. We encode each dialogue with \texttt{NV-Embed-v2} embeddings \cite{lee2025nvembed}. We then project the embeddings into a lower-dimensional space using UMAP \cite{mcinnes2018umap}, with \texttt{n\_components}=10 and \texttt{n\_neighbors}=15. Finally, we cluster the reduced representations with HDBSCAN \cite{malzer2020hybrid}. Since complaint issues vary in both frequency and density, we tune HDBSCAN by searching \texttt{min\_cluster\_size} over \(\{25,35,50,80\}\) and \texttt{min\_samples} over \(\{3,5,8,10\}\). For each topic, we review representative dialogues and topic keywords to identify the dominant complaint pattern and potential issue boundaries.

    \item \textbf{K-means-based subtopic exploration.}
    Because broad BERTopic clusters can contain multiple issue types, we further apply K-means clustering \cite{mcqueen1967some} within major topic clusters. We vary \(k\) from 5 to 25 and inspect representative samples from the resulting subclusters. This step helps uncover finer-grained distinctions within broad complaint themes.

    \item \textbf{Expert-guided coding and taxonomy development.}
    Using the BERTopic themes and K-means subclusters as analytic materials, we develop the final taxonomy through an expert-guided coding process. For each industry, the research team and hospitality and service domain professionals proceed in four steps. They define the major customer journey stages based on the industry service process and observed complaint contexts, review representative dialogues from the discovered topic groups and assign provisional complaint labels, compare labels within and across groups to merge overlapping labels and refine category boundaries, and retain recurring and interpretable categories relevant to complaint handling while removing or merging overly narrow and low-frequency issues. Each complaint category is then mapped to the corresponding journey stage.

    \item \textbf{LLM-based taxonomy assignment.}
    After finalizing the taxonomy, we use GPT-4o mini to classify the cleaned complaint dialogues into the finalized categories. Dialogues that cannot be assigned to any category are discarded.

\end{itemize}

\subsection{Scenario Data Provenance}
\label{app:scenario-provenance}

ServiceBench combines real complaint data with reconstructed context to support realistic multi-turn interactions. The source dialogues provide the original complaint facts, from which additional scenario components are reconstructed or generated. The selected scenarios are then manually reviewed and refined to improve realism and internal consistency while preserving the original facts. Table~\ref{tab:scenario-provenance} summarizes the provenance of each component.

\begin{table*}[t]
\centering
\scriptsize
\setlength{\tabcolsep}{5pt}
\renewcommand{\arraystretch}{1.15}
\begin{tabularx}{\textwidth}{C{0.17\textwidth}L{0.76\textwidth}}
\toprule
\textbf{Communication Style} & \textbf{Prompt Template} \\
\midrule

Assertive-Warm &
Maintain a clear, cooperative stance and focus on fair resolution. State the problem and what would feel fair, while still showing respect for the agent's effort. Give relevant details when needed and respond positively when the agent is helpful. Ask for clarification or a better next step when the response is vague or insufficient. \\

\midrule
Assertive-Cold &
Maintain a firm, skeptical stance and focus on concrete action. State the problem and expected outcome clearly, with little effort to sound friendly or cooperative. Treat vague explanations or unsupported offers as insufficient, and ask for accountability when the agent does not provide a credible path forward. Keep messages controlled but demanding. \\

\midrule
Passive-Warm &
Maintain a polite, patient, and cooperative stance. Use softer language and avoid direct confrontation. Share concerns gradually, with strong attention to whether the agent understands and treats the issue with care. Respond more openly when the agent shows understanding or provides a credible explanation. When the response feels unclear or insufficient, continue expressing concern gently rather than making strong demands. \\

\midrule
Passive-Cold &
Maintain a brief, withdrawn, and skeptical stance. Give limited details unless the agent asks specific, low-pressure follow-up questions. Show dissatisfaction through short replies or hesitation rather than direct confrontation. Become less responsive when the agent gives generic or repetitive responses. Engage more only when the agent shows that the concern is taken seriously and provides a credible path forward. \\

\bottomrule
\end{tabularx}
\caption{Communication style prompt.}
\label{tab:communication-style-prompts}
\end{table*}

\subsection{Complaint Case Quality}
\label{Quality Filtering}

We first filter the 14,373 candidate complaint cases using LLM-based judges. Each candidate is scored from 1 to 5 on coherence, plausibility, specificity, and category fit, and we retain only cases that receive at least 4 on all four dimensions. The candidates achieve average scores of 4.429 for coherence, 4.439 for plausibility, 4.441 for specificity, and 4.629 for category fit, with 86.68\% meeting the filtering threshold. From this qualified pool, we manually select 2,500 scenarios that most closely reflect realistic service situations and preserve the complaint facts and contextual information contained in the original dialogues. We then manually refine the selected cases to improve realism, completeness, and internal consistency before including them in the final benchmark.

\begin{itemize}

    \item \textbf{Coherence} assesses whether the complaint case is internally consistent without contradictions or unsupported event transitions.

    \item \textbf{Plausibility} assesses whether the complaint case is credible for the corresponding industry and realistic for complaint handling.

    \item \textbf{Specificity} assesses whether the case contains concrete details needed for complaint handling rather than generic descriptions.

    \item \textbf{Category fit} assesses whether the central service failure matches the assigned complaint category and customer journey stage.

\end{itemize}

\subsection{Customer Journey Maps}
\label{app:customer-journey-maps}

\begin{figure*}[t]
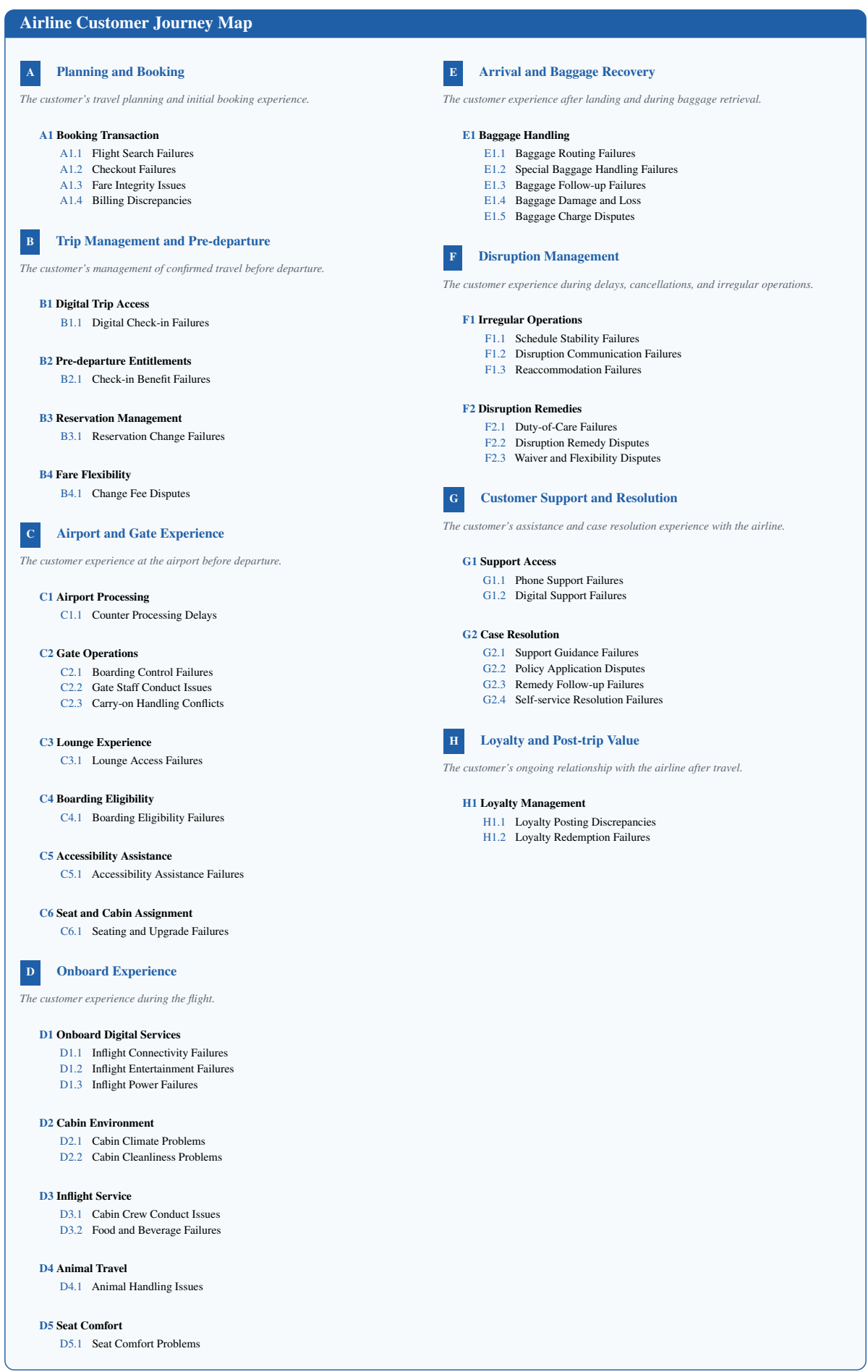

\centering
\setjourneycolor{airlineblue}
\begin{minipage}{0.985\textwidth}
\begin{journeymapbox}{Airline Customer Journey Map}

\begin{minipage}[t]{0.49\textwidth}
\phaseblock{A}{Planning and Booking}
{The customer’s travel planning and initial booking experience.}
\maincat{A1}{Booking Transaction}
\begin{compactsubcats}
  \subcat{A1.1}{Flight Search Failures}
  \subcat{A1.2}{Checkout Failures}
  \subcat{A1.3}{Fare Integrity Issues}
  \subcat{A1.4}{Billing Discrepancies}
\end{compactsubcats}

\phaseblock{B}{Trip Management and Pre-departure}
{The customer’s management of confirmed travel before departure.}
\maincat{B1}{Digital Trip Access}
\begin{compactsubcats}
  \subcat{B1.1}{Digital Check-in Failures}
\end{compactsubcats}
\maincat{B2}{Pre-departure Entitlements}
\begin{compactsubcats}
  \subcat{B2.1}{Check-in Benefit Failures}
\end{compactsubcats}
\maincat{B3}{Reservation Management}
\begin{compactsubcats}
  \subcat{B3.1}{Reservation Change Failures}
\end{compactsubcats}
\maincat{B4}{Fare Flexibility}
\begin{compactsubcats}
  \subcat{B4.1}{Change Fee Disputes}
\end{compactsubcats}

\phaseblock{C}{Airport and Gate Experience}
{The customer experience at the airport before departure.}
\maincat{C1}{Airport Processing}
\begin{compactsubcats}
  \subcat{C1.1}{Counter Processing Delays}
\end{compactsubcats}
\maincat{C2}{Gate Operations}
\begin{compactsubcats}
  \subcat{C2.1}{Boarding Control Failures}
  \subcat{C2.2}{Gate Staff Conduct Issues}
  \subcat{C2.3}{Carry-on Handling Conflicts}
\end{compactsubcats}
\maincat{C3}{Lounge Experience}
\begin{compactsubcats}
  \subcat{C3.1}{Lounge Access Failures}
\end{compactsubcats}
\maincat{C4}{Boarding Eligibility}
\begin{compactsubcats}
  \subcat{C4.1}{Boarding Eligibility Failures}
\end{compactsubcats}
\maincat{C5}{Accessibility Assistance}
\begin{compactsubcats}
  \subcat{C5.1}{Accessibility Assistance Failures}
\end{compactsubcats}
\maincat{C6}{Seat and Cabin Assignment}
\begin{compactsubcats}
  \subcat{C6.1}{Seating and Upgrade Failures}
\end{compactsubcats}

\phaseblock{D}{Onboard Experience}
{The customer experience during the flight.}
\maincat{D1}{Onboard Digital Services}
\begin{compactsubcats}
  \subcat{D1.1}{Inflight Connectivity Failures}
  \subcat{D1.2}{Inflight Entertainment Failures}
  \subcat{D1.3}{Inflight Power Failures}
\end{compactsubcats}
\maincat{D2}{Cabin Environment}
\begin{compactsubcats}
  \subcat{D2.1}{Cabin Climate Problems}
  \subcat{D2.2}{Cabin Cleanliness Problems}
\end{compactsubcats}
\maincat{D3}{Inflight Service}
\begin{compactsubcats}
  \subcat{D3.1}{Cabin Crew Conduct Issues}
  \subcat{D3.2}{Food and Beverage Failures}
\end{compactsubcats}
\maincat{D4}{Animal Travel}
\begin{compactsubcats}
  \subcat{D4.1}{Animal Handling Issues}
\end{compactsubcats}
\maincat{D5}{Seat Comfort}
\begin{compactsubcats}
  \subcat{D5.1}{Seat Comfort Problems}
\end{compactsubcats}
\end{minipage}
\hfill
\begin{minipage}[t]{0.49\textwidth}

\phaseblock{E}{Arrival and Baggage Recovery}
{The customer experience after landing and during baggage retrieval.}
\maincat{E1}{Baggage Handling}
\begin{compactsubcats}
  \subcat{E1.1}{Baggage Routing Failures}
  \subcat{E1.2}{Special Baggage Handling Failures}
  \subcat{E1.3}{Baggage Follow-up Failures}
  \subcat{E1.4}{Baggage Damage and Loss}
  \subcat{E1.5}{Baggage Charge Disputes}
\end{compactsubcats}

\phaseblock{F}{Disruption Management}
{The customer experience during delays, cancellations, and irregular operations.}
\maincat{F1}{Irregular Operations}
\begin{compactsubcats}
  \subcat{F1.1}{Schedule Stability Failures}
  \subcat{F1.2}{Disruption Communication Failures}
  \subcat{F1.3}{Reaccommodation Failures}
\end{compactsubcats}
\maincat{F2}{Disruption Remedies}
\begin{compactsubcats}
  \subcat{F2.1}{Duty-of-Care Failures}
  \subcat{F2.2}{Disruption Remedy Disputes}
  \subcat{F2.3}{Waiver and Flexibility Disputes}
\end{compactsubcats}

\phaseblock{G}{Customer Support and Resolution}
{The customer’s assistance and case resolution experience with the airline.}
\maincat{G1}{Support Access}
\begin{compactsubcats}
  \subcat{G1.1}{Phone Support Failures}
  \subcat{G1.2}{Digital Support Failures}
\end{compactsubcats}
\maincat{G2}{Case Resolution}
\begin{compactsubcats}
  \subcat{G2.1}{Support Guidance Failures}
  \subcat{G2.2}{Policy Application Disputes}
  \subcat{G2.3}{Remedy Follow-up Failures}
  \subcat{G2.4}{Self-service Resolution Failures}
\end{compactsubcats}

\phaseblock{H}{Loyalty and Post-trip Value}
{The customer’s ongoing relationship with the airline after travel.}
\maincat{H1}{Loyalty Management}
\begin{compactsubcats}
  \subcat{H1.1}{Loyalty Posting Discrepancies}
  \subcat{H1.2}{Loyalty Redemption Failures}
\end{compactsubcats}

\end{minipage}

\end{journeymapbox}
\end{minipage}
\caption{Overview of the airline customer journey map.}
\label{fig:airline-journey-map}
\end{figure*}

\begin{figure*}[t]
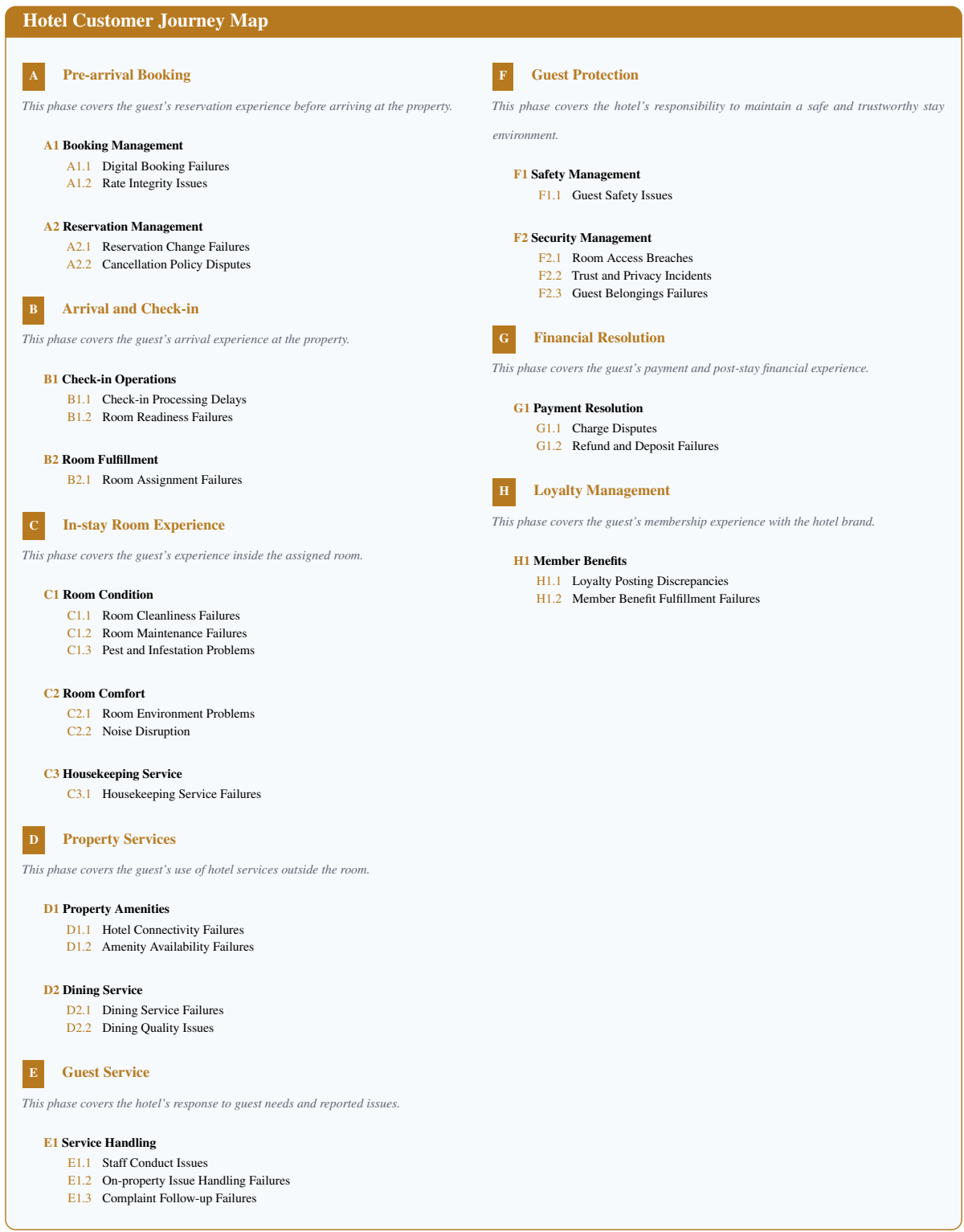

\centering
\setjourneycolor{hotelgold}
\begin{minipage}{0.985\textwidth}
\begin{journeymapbox}{Hotel Customer Journey Map}

\begin{minipage}[t]{0.49\textwidth}

\phaseblock{A}{Pre-arrival Booking}
{This phase covers the guest’s reservation experience before arriving at the property.}
\maincat{A1}{Booking Management}
\begin{compactsubcats}
  \subcat{A1.1}{Digital Booking Failures}
  \subcat{A1.2}{Rate Integrity Issues}
\end{compactsubcats}
\maincat{A2}{Reservation Management}
\begin{compactsubcats}
  \subcat{A2.1}{Reservation Change Failures}
  \subcat{A2.2}{Cancellation Policy Disputes}
\end{compactsubcats}

\phaseblock{B}{Arrival and Check-in}
{This phase covers the guest’s arrival experience at the property.}
\maincat{B1}{Check-in Operations}
\begin{compactsubcats}
  \subcat{B1.1}{Check-in Processing Delays}
  \subcat{B1.2}{Room Readiness Failures}
\end{compactsubcats}
\maincat{B2}{Room Fulfillment}
\begin{compactsubcats}
  \subcat{B2.1}{Room Assignment Failures}
\end{compactsubcats}

\phaseblock{C}{In-stay Room Experience}
{This phase covers the guest’s experience inside the assigned room.}
\maincat{C1}{Room Condition}
\begin{compactsubcats}
  \subcat{C1.1}{Room Cleanliness Failures}
  \subcat{C1.2}{Room Maintenance Failures}
  \subcat{C1.3}{Pest and Infestation Problems}
\end{compactsubcats}
\maincat{C2}{Room Comfort}
\begin{compactsubcats}
  \subcat{C2.1}{Room Environment Problems}
  \subcat{C2.2}{Noise Disruption}
\end{compactsubcats}
\maincat{C3}{Housekeeping Service}
\begin{compactsubcats}
  \subcat{C3.1}{Housekeeping Service Failures}
\end{compactsubcats}

\phaseblock{D}{Property Services}
{This phase covers the guest’s use of hotel services outside the room.}
\maincat{D1}{Property Amenities}
\begin{compactsubcats}
  \subcat{D1.1}{Hotel Connectivity Failures}
  \subcat{D1.2}{Amenity Availability Failures}
\end{compactsubcats}
\maincat{D2}{Dining Service}
\begin{compactsubcats}
  \subcat{D2.1}{Dining Service Failures}
  \subcat{D2.2}{Dining Quality Issues}
\end{compactsubcats}

\phaseblock{E}{Guest Service}
{This phase covers the hotel’s response to guest needs and reported issues.}
\maincat{E1}{Service Handling}
\begin{compactsubcats}
  \subcat{E1.1}{Staff Conduct Issues}
  \subcat{E1.2}{On-property Issue Handling Failures}
  \subcat{E1.3}{Complaint Follow-up Failures}
\end{compactsubcats}

\end{minipage}
\hfill
\begin{minipage}[t]{0.49\textwidth}

\phaseblock{F}{Guest Protection}
{This phase covers the hotel’s responsibility to maintain a safe and trustworthy stay environment.}
\maincat{F1}{Safety Management}
\begin{compactsubcats}
  \subcat{F1.1}{Guest Safety Issues}
\end{compactsubcats}
\maincat{F2}{Security Management}
\begin{compactsubcats}
  \subcat{F2.1}{Room Access Breaches}
  \subcat{F2.2}{Trust and Privacy Incidents}
  \subcat{F2.3}{Guest Belongings Failures}
\end{compactsubcats}

\phaseblock{G}{Financial Resolution}
{This phase covers the guest’s payment and post-stay financial experience.}
\maincat{G1}{Payment Resolution}
\begin{compactsubcats}
  \subcat{G1.1}{Charge Disputes}
  \subcat{G1.2}{Refund and Deposit Failures}
\end{compactsubcats}

\phaseblock{H}{Loyalty Management}
{This phase covers the guest’s membership experience with the hotel brand.}
\maincat{H1}{Member Benefits}
\begin{compactsubcats}
  \subcat{H1.1}{Loyalty Posting Discrepancies}
  \subcat{H1.2}{Member Benefit Fulfillment Failures}
\end{compactsubcats}

\end{minipage}

\end{journeymapbox}
\end{minipage}
\caption{Overview of the hotel customer journey map.}
\label{fig:hotel-journey-map}
\end{figure*}

\begin{figure*}[t]
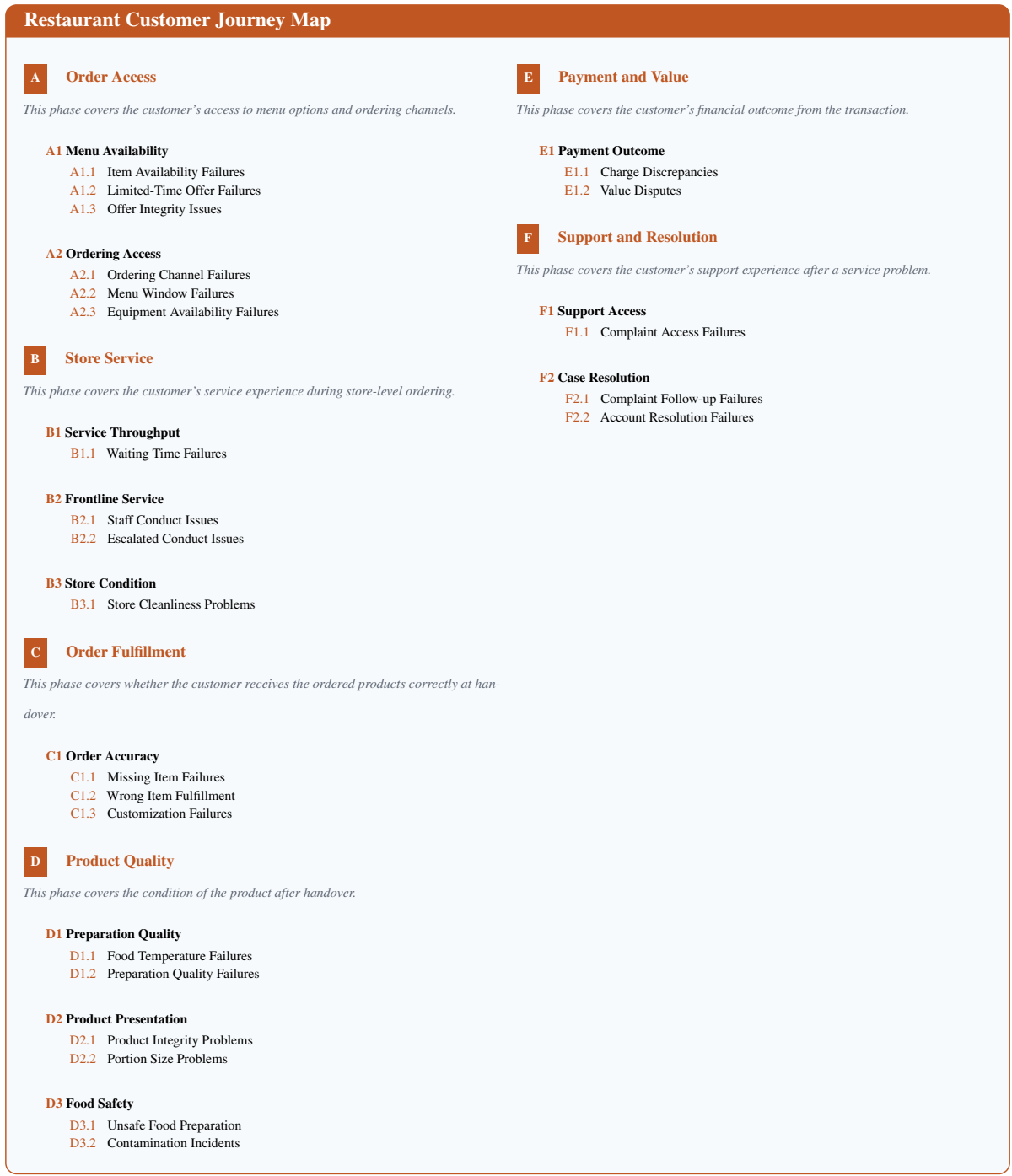

\centering
\setjourneycolor{restaurantorange}
\begin{minipage}{0.985\textwidth}
\begin{journeymapbox}{Restaurant Customer Journey Map}

\begin{minipage}[t]{0.49\textwidth}

\phaseblock{A}{Order Access}
{This phase covers the customer’s access to menu options and ordering channels.}
\maincat{A1}{Menu Availability}
\begin{compactsubcats}
  \subcat{A1.1}{Item Availability Failures}
  \subcat{A1.2}{Limited-Time Offer Failures}
  \subcat{A1.3}{Offer Integrity Issues}
\end{compactsubcats}
\maincat{A2}{Ordering Access}
\begin{compactsubcats}
  \subcat{A2.1}{Ordering Channel Failures}
  \subcat{A2.2}{Menu Window Failures}
  \subcat{A2.3}{Equipment Availability Failures}
\end{compactsubcats}

\phaseblock{B}{Store Service}
{This phase covers the customer’s service experience during store-level ordering.}
\maincat{B1}{Service Throughput}
\begin{compactsubcats}
  \subcat{B1.1}{Waiting Time Failures}
\end{compactsubcats}
\maincat{B2}{Frontline Service}
\begin{compactsubcats}
  \subcat{B2.1}{Staff Conduct Issues}
  \subcat{B2.2}{Escalated Conduct Issues}
\end{compactsubcats}
\maincat{B3}{Store Condition}
\begin{compactsubcats}
  \subcat{B3.1}{Store Cleanliness Problems}
\end{compactsubcats}

\phaseblock{C}{Order Fulfillment}
{This phase covers whether the customer receives the ordered products correctly at handover.}
\maincat{C1}{Order Accuracy}
\begin{compactsubcats}
  \subcat{C1.1}{Missing Item Failures}
  \subcat{C1.2}{Wrong Item Fulfillment}
  \subcat{C1.3}{Customization Failures}
\end{compactsubcats}

\phaseblock{D}{Product Quality}
{This phase covers the condition of the product after handover.}
\maincat{D1}{Preparation Quality}
\begin{compactsubcats}
  \subcat{D1.1}{Food Temperature Failures}
  \subcat{D1.2}{Preparation Quality Failures}
\end{compactsubcats}
\maincat{D2}{Product Presentation}
\begin{compactsubcats}
  \subcat{D2.1}{Product Integrity Problems}
  \subcat{D2.2}{Portion Size Problems}
\end{compactsubcats}
\maincat{D3}{Food Safety}
\begin{compactsubcats}
  \subcat{D3.1}{Unsafe Food Preparation}
  \subcat{D3.2}{Contamination Incidents}
\end{compactsubcats}

\end{minipage}
\hfill
\begin{minipage}[t]{0.49\textwidth}

\phaseblock{E}{Payment and Value}
{This phase covers the customer’s financial outcome from the transaction.}
\maincat{E1}{Payment Outcome}
\begin{compactsubcats}
  \subcat{E1.1}{Charge Discrepancies}
  \subcat{E1.2}{Value Disputes}
\end{compactsubcats}

\phaseblock{F}{Support and Resolution}
{This phase covers the customer’s support experience after a service problem.}
\maincat{F1}{Support Access}
\begin{compactsubcats}
  \subcat{F1.1}{Complaint Access Failures}
\end{compactsubcats}
\maincat{F2}{Case Resolution}
\begin{compactsubcats}
  \subcat{F2.1}{Complaint Follow-up Failures}
  \subcat{F2.2}{Account Resolution Failures}
\end{compactsubcats}

\end{minipage}

\end{journeymapbox}
\end{minipage}
\caption{Overview of the restaurant customer journey map.}
\label{fig:restaurant-journey-map}
\end{figure*}

\begin{figure*}[t]
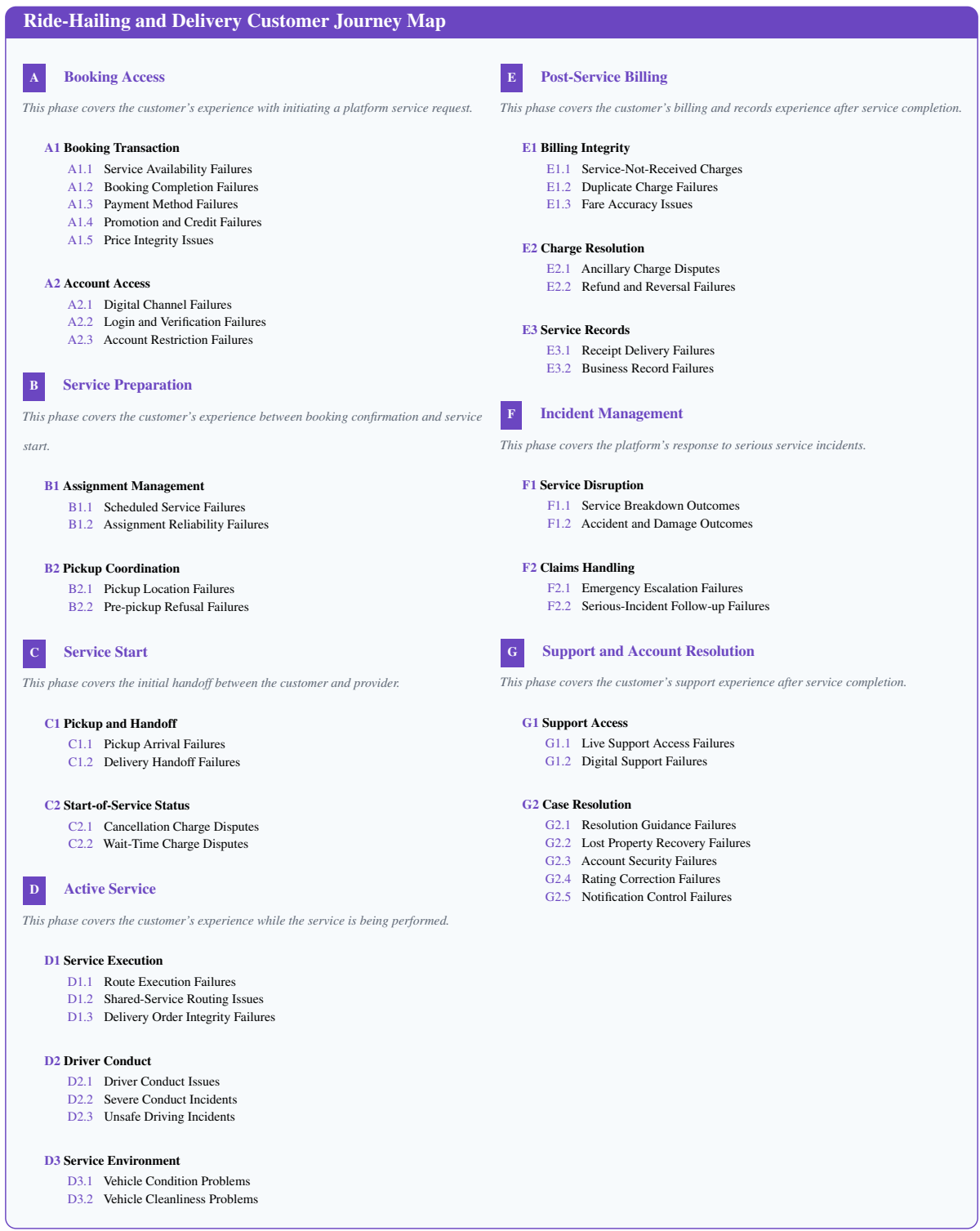

\centering
\setjourneycolor{platformpurple}
\begin{minipage}{0.985\textwidth}
\begin{journeymapbox}{Ride-Hailing and Delivery Customer Journey Map}

\begin{minipage}[t]{0.49\textwidth}

\phaseblock{A}{Booking Access}
{This phase covers the customer’s experience with initiating a platform service request.}
\maincat{A1}{Booking Transaction}
\begin{compactsubcats}
  \subcat{A1.1}{Service Availability Failures}
  \subcat{A1.2}{Booking Completion Failures}
  \subcat{A1.3}{Payment Method Failures}
  \subcat{A1.4}{Promotion and Credit Failures}
  \subcat{A1.5}{Price Integrity Issues}
\end{compactsubcats}
\maincat{A2}{Account Access}
\begin{compactsubcats}
  \subcat{A2.1}{Digital Channel Failures}
  \subcat{A2.2}{Login and Verification Failures}
  \subcat{A2.3}{Account Restriction Failures}
\end{compactsubcats}

\phaseblock{B}{Service Preparation}
{This phase covers the customer’s experience between booking confirmation and service start.}
\maincat{B1}{Assignment Management}
\begin{compactsubcats}
  \subcat{B1.1}{Scheduled Service Failures}
  \subcat{B1.2}{Assignment Reliability Failures}
\end{compactsubcats}
\maincat{B2}{Pickup Coordination}
\begin{compactsubcats}
  \subcat{B2.1}{Pickup Location Failures}
  \subcat{B2.2}{Pre-pickup Refusal Failures}
\end{compactsubcats}

\phaseblock{C}{Service Start}
{This phase covers the initial handoff between the customer and provider.}
\maincat{C1}{Pickup and Handoff}
\begin{compactsubcats}
  \subcat{C1.1}{Pickup Arrival Failures}
  \subcat{C1.2}{Delivery Handoff Failures}
\end{compactsubcats}
\maincat{C2}{Start-of-Service Status}
\begin{compactsubcats}
  \subcat{C2.1}{Cancellation Charge Disputes}
  \subcat{C2.2}{Wait-Time Charge Disputes}
\end{compactsubcats}

\phaseblock{D}{Active Service}
{This phase covers the customer’s experience while the service is being performed.}
\maincat{D1}{Service Execution}
\begin{compactsubcats}
  \subcat{D1.1}{Route Execution Failures}
  \subcat{D1.2}{Shared-Service Routing Issues}
  \subcat{D1.3}{Delivery Order Integrity Failures}
\end{compactsubcats}
\maincat{D2}{Driver Conduct}
\begin{compactsubcats}
  \subcat{D2.1}{Driver Conduct Issues}
  \subcat{D2.2}{Severe Conduct Incidents}
  \subcat{D2.3}{Unsafe Driving Incidents}
\end{compactsubcats}
\maincat{D3}{Service Environment}
\begin{compactsubcats}
  \subcat{D3.1}{Vehicle Condition Problems}
  \subcat{D3.2}{Vehicle Cleanliness Problems}
\end{compactsubcats}

\end{minipage}
\hfill
\begin{minipage}[t]{0.49\textwidth}

\phaseblock{E}{Post-Service Billing}
{This phase covers the customer’s billing and records experience after service completion.}
\maincat{E1}{Billing Integrity}
\begin{compactsubcats}
  \subcat{E1.1}{Service-Not-Received Charges}
  \subcat{E1.2}{Duplicate Charge Failures}
  \subcat{E1.3}{Fare Accuracy Issues}
\end{compactsubcats}
\maincat{E2}{Charge Resolution}
\begin{compactsubcats}
  \subcat{E2.1}{Ancillary Charge Disputes}
  \subcat{E2.2}{Refund and Reversal Failures}
\end{compactsubcats}
\maincat{E3}{Service Records}
\begin{compactsubcats}
  \subcat{E3.1}{Receipt Delivery Failures}
  \subcat{E3.2}{Business Record Failures}
\end{compactsubcats}

\phaseblock{F}{Incident Management}
{This phase covers the platform’s response to serious service incidents.}
\maincat{F1}{Service Disruption}
\begin{compactsubcats}
  \subcat{F1.1}{Service Breakdown Outcomes}
  \subcat{F1.2}{Accident and Damage Outcomes}
\end{compactsubcats}
\maincat{F2}{Claims Handling}
\begin{compactsubcats}
  \subcat{F2.1}{Emergency Escalation Failures}
  \subcat{F2.2}{Serious-Incident Follow-up Failures}
\end{compactsubcats}

\phaseblock{G}{Support and Account Resolution}
{This phase covers the customer’s support experience after service completion.}
\maincat{G1}{Support Access}
\begin{compactsubcats}
  \subcat{G1.1}{Live Support Access Failures}
  \subcat{G1.2}{Digital Support Failures}
\end{compactsubcats}
\maincat{G2}{Case Resolution}
\begin{compactsubcats}
  \subcat{G2.1}{Resolution Guidance Failures}
  \subcat{G2.2}{Lost Property Recovery Failures}
  \subcat{G2.3}{Account Security Failures}
  \subcat{G2.4}{Rating Correction Failures}
  \subcat{G2.5}{Notification Control Failures}
\end{compactsubcats}

\end{minipage}

\end{journeymapbox}
\end{minipage}
\caption{Overview of the ride-hailing and delivery customer journey map.}
\label{fig:mobility-journey-map}
\end{figure*}

The detailed industry-level customer journey maps developed for \textsc{ServiceBench} are provided in Figures~\ref{fig:airline-journey-map}, \ref{fig:hotel-journey-map}, \ref{fig:restaurant-journey-map}, \ref{fig:retail-journey-map}, and \ref{fig:mobility-journey-map}. These figures present the journey stages and complaint categories used to organize benchmark coverage across the five service industries.

\section{Implementation Details}
\label{app:implementation-details}

We select seven service agent models from OpenAI, Google, DeepSeek, Meta, and Qwen to cover diverse model families, capability levels, and both proprietary and open weight models. This selection provides broad coverage of contemporary LLM capabilities while keeping large scale multi-turn evaluation computationally feasible.

For customer simulation, personas are sampled from Nemotron-Personas \cite{nvidia/Nemotron-Personas-USA}, while four communication styles are developed based on the Interpersonal Circumplex framework \cite{wiggins1991agency}, varying in agency and communion. To reflect realistic complaint handling practices, we work with hospitality and service professionals to develop representative prior complaint cases, standard operating procedures (SOPs), and material remedy rules. The prior cases provide examples of complaint handling decisions, while the SOPs offer procedural guidance for resolution. The material remedy rules distinguish discretionary goodwill, refunds, and compensation, with a standardized goodwill limit of USD 200 per case to approximate frontline service authority and ensure comparability across industries.

We use consistent model and decoding settings across experiments. The customer simulator uses GPT-5 and DeepSeek-V4-Pro with temperature $0.7$; the corresponding prompts are shown in Figures~\ref{fig:prompt-customer-initial} and \ref{fig:prompt-customer-subsequent}, and the communication style templates in Table~\ref{tab:communication-style-prompts}. All service agent models use temperature $0.2$ to balance response stability and conversational naturalness, with the common prompt shown in Figure~\ref{fig:prompt-service-agent}. We adopt GPT-5 as the primary justice evaluator and use DeepSeek-V4-Pro, Llama 3.3 70B, and Qwen3-32B as additional judges. All four judges use temperature $0$, with the evaluation prompt shown in Figure~\ref{fig:prompt-justice-evaluation}. The outcome extractor uses GPT-4o mini with temperature $0$; its prompt is shown in Figure~\ref{fig:prompt-outcome-extraction}.

\subsection{Prompt Templates}
\label{app:prompt-templates}

The communication style instructions used to condition the customer simulator are reported in Table~\ref{tab:communication-style-prompts}. The customer simulator prompts for the initial complaint and subsequent interaction turns are provided in Figures~\ref{fig:prompt-customer-initial} and~\ref{fig:prompt-customer-subsequent}. The common service agent prompt is shown in Figure~\ref{fig:prompt-service-agent}. The prompts used for justice evaluation and dialogue-level outcome extraction are provided in Figures~\ref{fig:prompt-justice-evaluation} and~\ref{fig:prompt-outcome-extraction}.

\subsection{Post-Training Details} 
\label{app:post-training} 
 
We fine-tune Qwen3-8B using supervised fine-tuning (SFT) followed by direct preference optimization (DPO). We first split the scenarios into separate training and evaluation sets, balanced across service industries and customer journey stages. The SFT data are constructed from service interaction rollouts generated by the seven models evaluated in our main experiments. We select high quality rollouts with scores of at least 3 on each justice dimension. Each rollout is converted into turn level examples using the preceding dialogue context and the corresponding service response. We then manually screen these examples and retain only high quality training instances, resulting in 5,125 SFT examples. SFT is conducted for two epochs with a learning rate of $3\times10^{-4}$. For DPO, the trained SFT model generates five candidate responses for each selected turn state. We then manually select preference pairs that provide useful training signals for the behavioral limitations identified in our earlier analyses, resulting in 1,018 chosen--rejected pairs. Starting from the SFT model, DPO is conducted for two epochs with a learning rate of $5\times10^{-6}$ and $\beta=0.1$.

\clearpage

\begin{figure*}[t]
\centering
\setjourneycolor{retailgreen}
\begin{minipage}{0.985\textwidth}
\begin{journeymapbox}{Retail Customer Journey Map}

\begin{minipage}[t]{0.49\textwidth}

\phaseblock{A}{Product Discovery}
{This phase covers the customer’s product search and availability assessment before purchase.}
\maincat{A1}{Inventory Visibility}
\begin{compactsubcats}
  \subcat{A1.1}{Inventory Accuracy Failures}
  \subcat{A1.2}{Stock Availability Failures}
  \subcat{A1.3}{Assortment Gaps}
\end{compactsubcats}
\maincat{A2}{Limited-Item Access}
\begin{compactsubcats}
  \subcat{A2.1}{Launch Availability Failures}
  \subcat{A2.2}{Limited-Item Access Failures}
  \subcat{A2.3}{Sell-Block Failures}
\end{compactsubcats}

\phaseblock{B}{Price and Promotion}
{This phase covers the customer’s evaluation of retail offer value.}
\maincat{B1}{Offer Integrity}
\begin{compactsubcats}
  \subcat{B1.1}{Promotion Clarity Problems}
  \subcat{B1.2}{Coupon Redemption Failures}
  \subcat{B1.3}{Promotional Benefit Failures}
\end{compactsubcats}
\maincat{B2}{Price Integrity}
\begin{compactsubcats}
  \subcat{B2.1}{Price Match Disputes}
  \subcat{B2.2}{Store Pricing Errors}
\end{compactsubcats}

\phaseblock{C}{Digital Shopping}
{This phase covers the customer’s experience with the retailer’s online purchase channel.}
\maincat{C1}{Digital Access}
\begin{compactsubcats}
  \subcat{C1.1}{Website and App Failures}
  \subcat{C1.2}{Account Access Failures}
  \subcat{C1.3}{Digital Usability Problems}
\end{compactsubcats}
\maincat{C2}{Online Checkout}
\begin{compactsubcats}
  \subcat{C2.1}{Checkout Completion Failures}
  \subcat{C2.2}{Checkout Payment Failures}
\end{compactsubcats}

\phaseblock{D}{Store Shopping}
{This phase covers the customer’s experience within the physical retail store.}
\maincat{D1}{Store Environment}
\begin{compactsubcats}
  \subcat{D1.1}{Store Navigation Problems}
  \subcat{D1.2}{Cart Availability Problems}
  \subcat{D1.3}{Store Cleanliness Problems}
  \subcat{D1.4}{Store Hours Accuracy Problems}
\end{compactsubcats}
\maincat{D2}{Store Service}
\begin{compactsubcats}
  \subcat{D2.1}{Specialist Service Delays}
  \subcat{D2.2}{Checkout Capacity Failures}
  \subcat{D2.3}{Staff Conduct Issues}
  \subcat{D2.4}{Discriminatory Service Incidents}
\end{compactsubcats}

\phaseblock{E}{Fulfillment}
{This phase covers order completion after purchase.}
\maincat{E1}{Pickup Fulfillment}
\begin{compactsubcats}
  \subcat{E1.1}{Pickup Execution Failures}
  \subcat{E1.2}{Pre-order Delay Failures}
  \subcat{E1.3}{Pre-order Cancellation Failures}
\end{compactsubcats}
\maincat{E2}{Delivery Fulfillment}
\begin{compactsubcats}
  \subcat{E2.1}{Shipping Visibility Failures}
  \subcat{E2.2}{Delivery Handling Failures}
  \subcat{E2.3}{Wrong Item Fulfillment}
  \subcat{E2.4}{Incomplete Delivery}
\end{compactsubcats}

\end{minipage}
\hfill
\begin{minipage}[t]{0.49\textwidth}

\phaseblock{F}{Product Experience}
{This phase covers the customer’s experience with the purchased product.}
\maincat{F1}{Product Condition}
\begin{compactsubcats}
  \subcat{F1.1}{Early Product Failure}
  \subcat{F1.2}{Tampered Product Condition}
  \subcat{F1.3}{Product Mismatch}
\end{compactsubcats}
\maincat{F2}{Consumable Quality}
\begin{compactsubcats}
  \subcat{F2.1}{Unsafe Consumables}
\end{compactsubcats}
\maincat{F3}{Product Safety}
\begin{compactsubcats}
  \subcat{F3.1}{Product Hazard Concerns}
\end{compactsubcats}

\phaseblock{G}{Returns and Resolution}
{This phase covers the retailer’s handling of post-purchase problems.}
\maincat{G1}{Refund Resolution}
\begin{compactsubcats}
  \subcat{G1.1}{Refund Execution Failures}
  \subcat{G1.2}{Order Reversal Problems}
\end{compactsubcats}
\maincat{G2}{Return Handling}
\begin{compactsubcats}
  \subcat{G2.1}{Return Friction}
  \subcat{G2.2}{Replacement Execution Failures}
\end{compactsubcats}
\maincat{G3}{Customer Support}
\begin{compactsubcats}
  \subcat{G3.1}{Contact Center Access Failures}
  \subcat{G3.2}{Case Follow-up Failures}
  \subcat{G3.3}{Support Interaction Failures}
\end{compactsubcats}

\phaseblock{H}{Account and Relationship Management}
{This phase covers the customer’s ongoing relationship with the retailer.}
\maincat{H1}{Retail Account Services}
\begin{compactsubcats}
  \subcat{H1.1}{Store Card Account Problems}
\end{compactsubcats}
\maincat{H2}{Registry Services}
\begin{compactsubcats}
  \subcat{H2.1}{Registry Fulfillment Failures}
\end{compactsubcats}
\maincat{H3}{Communication Preferences}
\begin{compactsubcats}
  \subcat{H3.1}{Marketing Preference Problems}
\end{compactsubcats}

\end{minipage}

\end{journeymapbox}
\end{minipage}
\caption{Overview of the retail customer journey map.}
\label{fig:retail-journey-map}
\end{figure*}

\begin{figure*}[p]
\begin{promptlisting}{Customer Agent Prompt (Initial Complaint Message)}
You are role playing as a dissatisfied customer. A service agent has contacted you after you filed a formal complaint about a poor service experience with the company. This is your first conversation session with the service agent. Any clear promise made by the service agent will be treated as a company commitment. Your task is to generate a realistic and appropriate initial message that the customer might say at the start of the interaction.

# Guidelines for Your Role as the Customer:
1. Stay consistent with the provided customer profile, communication style, and experience background.
2. Your initial message should express disappointment or anger about the experience.
3. Share your experience and feelings gradually, as you would in a natural conversation, rather than providing everything at once.
4. Your goal is to communicate dissatisfaction clearly and seek a fair response from the company.

# Customer Profile:
{customer_profile}

# Communication Style:
{communication_style_prompt}

# Experience Background:
Service context: {service_context}
Experience: {experience_background}

# Task:
Write the customer's initial message. The message should reflect your current level of dissatisfaction and anger about the poor service encounter.

## Thought Process
Before generating your initial message, think through the situation as the customer. Consider:
- What happened in the service experience and why it made you dissatisfied.
- Your current emotional state.
- How your customer profile and communication style shape the way you complain.
- What you want the company to understand first.

## Message Generation
Based on your thought process, generate:
A natural, human-like chat message that expresses your dissatisfaction and anger about the poor service encounter.

# Output Format:
Thought: [Your reasoning about what to say to the service agent]

Message: [Your initial message for the service agent, 10-50 words]

# Note:
- The message should be suitable for online contact: natural and conversational.
- Keep your message 10-50 words.

Stay in character as the complaining customer throughout your output, following the above guidelines carefully.
\end{promptlisting}
\caption{Customer agent prompt (initial complaint message).}
\label{fig:prompt-customer-initial}
\end{figure*}

\begin{figure*}[p]
\begin{promptlisting}{Customer Agent Prompt}
You are role playing as a dissatisfied customer. A service agent has contacted you after you filed a formal complaint about a poor service experience with the company. This is your first conversation session with the service agent. Any clear promise made by the service agent will be treated as a company commitment. Your task is to generate a realistic and appropriate response that the customer might make when expressing dissatisfaction and seeking fair resolution.

# Guidelines for Your Role as the Customer:
1. Stay consistent with the provided customer profile, communication style, and experience background.
2. Use the resolution expectation as guidance for what you initially consider fair, but do not treat it as a rigid goal.
3. Respond naturally to the service agent, showing whether the reply addresses your concern or changes your level of satisfaction.
4. Share your experience, feelings, and expectations gradually, as you would in a natural conversation, rather than providing everything at once.

# Customer Profile:
{customer_profile}

# Communication Style:
{communication_style_prompt}

# Experience Background:
Service context: {service_context}
Experience: {experience_background}

# Expectation:
{customer_expectation}

# Conversation History:
{dialogue_history}

# Task:
Use the conversation history to generate the next response you would give to the service agent. The response should follow naturally and reflect your current reaction and expectation.

## Thought Process
Before generating your next message, think through the situation as the customer. Consider:
- What the service agent has most recently said or offered.
- Whether the response addresses your concern and expectation.
- Whether the response has persuaded you or changed your expectation.
- What would help move the issue toward a fair resolution.
- How your customer profile and communication style shape your response.

## Message Generation
Based on your thought process, generate a natural, human-like chat message that expresses your current level of satisfaction or dissatisfaction and expectation.

If any of the following conditions are met, output only "terminate conversation":
1. The service agent has adequately addressed your expectation.
2. Even if your expectation is not fully met, you are persuaded that the response, proposed next step, or resolution is fair and reasonable.
3. The conversation is getting repetitive or no longer productive.

# Output Format:
Thought: [Your analysis of the current situation, whether to terminate, and what you want to say]

Message: [Your next message to the service agent, 1-50 words]

# Note:
- The message should be suitable for online contact: natural and conversational.
- Keep your message 1-50 words.
- Do not repeatedly ask for confirmation once a clear commitment has been given.

Stay in character as the complaining customer throughout your output, following the above guidelines carefully.
\end{promptlisting}
\caption{Customer agent prompt.}
\label{fig:prompt-customer-subsequent}
\end{figure*}

\begin{figure*}[p]
\begin{promptlisting}{Service Agent Prompt}
You are a service agent from Customer Relations handling a formal complaint case from a dissatisfied customer. This is your first conversation session with the customer. Any promise you make will be treated as a company commitment. Your task is to generate a realistic and appropriate response that a service agent might give when trying to address the customer's dissatisfaction and support retention.

# Guidelines for Your Role as the Service Agent:
1. Ask relevant questions when needed, understand the customer's situation, and provide only solutions that are immediately available.
2. Use two recovery strategies when appropriate: symbolic recovery, which involves communicative and interpersonal responses such as apology and empathy, and utilitarian recovery, which involves material remedies.
3. Use the casebook reference as general guidance for similar complaints and remedy calibration, but do not treat it as a strict template. Tailor your response to the specific situation, customer, and conversation flow.
4. Your response will be evaluated on whether it is fair and appropriate, avoids unjustified remedy promises, and uses the minimum necessary remedy to address the issue.

# Guidelines for Material Remedy:
1. You can ONLY offer goodwill, refund, and compensation as material remedies.
2. Goodwill is a gesture offered to express apology and preserve customer goodwill. You may offer a goodwill voucher or credit worth up to USD 200 per case. Offer goodwill only when the issue is a company-side issue rather than a service limitation or unavoidable circumstance, or when it has caused significant customer impact.
3. A refund is reimbursement of the amount paid by the customer for the product or service. You may offer a refund up to the amount paid by the customer. Offer a refund only when the service was not delivered as promised, the product was defective, or the issue was severe enough to significantly affect the promised service experience.
4. Compensation is reimbursement for loss caused by the service failure. You may offer compensation up to the amount claimed by the customer. Offer compensation only when the issue caused significant customer loss and was due to a company-side issue.
5. Refunds and compensation should be promised only when supported by the record. If there is uncertainty, you may only promise that the issue will be reviewed later for further investigation.

# Company Description:
{company_description}

# Case Packet:
Complaint category: {category_name}
Complaint summary: {complaint_summary}
Available record: {company_record}

# Casebook Reference:
{case_reference_customer_issue}
{case_reference_previous_handling}

# Conversation History:
{dialogue_history}

# Task:
Use the conversation history to generate the next response you would give to the customer.

## Thought Process
Before generating your response, think through the situation as the service agent. Consider:
- The customer's main concern, emotion, and expected outcome.
- Whether more information is needed before offering a resolution.
- What is confirmed by the available record and what requires follow-up.
- The appropriate recovery strategy: symbolic recovery, utilitarian recovery, or both.
- Whether a material remedy is justified and, if so, the lowest suitable level.

## Message Generation
Based on your thought process, generate:
A natural chat message to the customer that addresses their dissatisfaction and supports retention.

# Output Format:
Thought: [Your analysis of the current situation and how you want to respond]

Message: [Your next message to the customer]

# Note:
- The message should be in a natural, professional online messaging style, not a formal email or letter.
- Keep the message 20-80 words.
- A fair resolution does not require satisfying every customer demand, especially when the request is unreasonable or unsupported.
- If goodwill is justified, start with the lowest reasonable amount and be conservative when offering higher amounts.
- Remain proactive in addressing the issue.

\end{promptlisting}
\caption{Service agent prompt.}
\label{fig:prompt-service-agent}
\end{figure*}

\begin{figure*}[p]
\begin{promptlisting}{Justice Evaluation Prompt}
You are a strict complaint handling evaluation judge. Your task is to evaluate the quality of the service agent's handling of a completed customer complaint dialogue. Be conservative and selective in scoring.

# Task

Evaluate the completed dialogue on three justice dimensions:
1. Interactional justice
2. Procedural justice
3. Distributive justice

For each dimension:
1. Think carefully about the main strengths and weaknesses.
2. Assign one final score from 1 to 5.

# Scoring Dimensions
## 1. Interactional Justice

Interactional justice evaluates the fairness and quality of the agent's interpersonal treatment across the completed dialogue. A strong response addresses the customer's concern and emotional state in a specific, attentive way rather than relying on generic service language. Follow-up questions should be rewarded only when they help the customer feel heard and move the conversation forward. The score should be lower when the agent repeats available information, delays progress, or makes the dialogue feel circular, dismissive, or unproductive. Do not reward politeness, fluency, or generosity by themselves.

## 2. Procedural Justice

Procedural justice evaluates whether the complaint handling process is clear, transparent, and credible. A strong response uses the available case information to explain what is known, what remains uncertain, and what can realistically happen next. The agent should provide a credible path for review, coordination, or remedy processing when the issue cannot be resolved immediately. The score should be lower when the agent relies on vague escalation, unrealistic timelines, incorrect coordination, or repeated promises of future review without meaningful process clarity.

## 3. Distributive Justice

Distributive justice evaluates whether the outcome is fair and proportionate to the service issue, customer impact, and company responsibility. A strong response provides an appropriate communicative response and uses material remedies only when they are justified by the case. The score should reflect whether the agent reaches a reasonable outcome, not whether the customer's full demand is satisfied. Refusing excessive or unsupported demands should not be penalized. A response without material remedy can still score well when the agent provides a fair explanation, review process, or practical guidance. The score should be lower when the outcome under-recovers serious harm or violates the remedy rules.

# Agent Material Remedy Rule

Use the following material remedy rules when evaluating distributive justice:
{material_remedy_rule}

# Output Format

Return valid JSON only. Do not include any text outside the JSON.

{
  "interactional_justice": {
    "score": <1-5>
  },
  "procedural_justice": {
    "score": <1-5>
  },
  "distributive_justice": {
    "score": <1-5>
  }
}

# Case to Evaluate
{Case}
\end{promptlisting}
\caption{Justice evaluation prompt.}
\label{fig:prompt-justice-evaluation}
\end{figure*}

\begin{figure*}[p]
\begin{promptlisting}{Outcome Extraction Prompt}
You are a structured extraction assistant. Read the completed customer service dialogue and extract three dialogue level outcomes: Disengagement, Agreement, and Goodwill Value.

# Extraction Rules

## 1. Disengagement

Disengagement captures whether the customer leaves because the agent's handling remains unacceptable or the conversation is no longer progressing. Use the last turn customer thought as supporting evidence of the customer's final assessment.

## 2. Agreement

Agreement captures whether the customer accepts the current handling and stops seeking further action in the dialogue. This can include acceptance of a remedy or a reasonable next step.

## 3. Goodwill Value

Goodwill Value captures the final specific USD amount of goodwill, voucher, or credit offered by the agent. Count only amounts stated by the agent, and use the highest amount that is offered or promised. Do not count amounts that are only under review, and do not count refunds, compensation, reimbursements, or other billing adjustments as goodwill. If no specific USD goodwill amount is offered, return 0.

# Notes

Disengagement and Agreement should not both be "yes" for the same case. If the customer neither clearly disengages nor clearly accepts the current handling, set both as "no".

# Output Format

Return valid JSON only. Do not include any text outside the JSON.

{
  "disengagement": "<yes/no>",
  "agreement": "<yes/no>",
  "goodwill_value": <number>
}

# Case to Extract
{Case}
\end{promptlisting}
\caption{Outcome extraction prompt.}
\label{fig:prompt-outcome-extraction}
\end{figure*}

\begin{figure*}[t]
\centering
\begin{casebox}{Case Study 1}
\scriptsize

\begin{casebackground}
\textbf{Category:} Driver Conduct Issues

\textbf{Complaint summary:} The customer reports that an accessible transport service failed to provide required assistance and that staff conduct was dismissive.
\end{casebackground}

\vspace{0.5em}

\begin{dialogueturn}{Turn 1}
\textbf{Customer:} I booked an accessible ride from SFO, and your driver refused to help with my wheelchair and made comments the whole way. I’m 71, and this felt unsafe and humiliating. What’s being done about that driver, and how are you making this right?

\vspace{0.35em}
\textbf{Agent:} I’m sorry you experienced this, Mr. Leyva. That kind of treatment is unacceptable, especially on an accessible ride. Your safety and dignity matter to us. To make sure we address this with our Safety and Ride Operations teams, could you share a bit more about what the driver said or did when you needed assistance?
\end{dialogueturn}

\vspace{0.45em}

\begin{dialogueturn}{Turn 2}
\textbf{Customer:} He pulled up, saw my wheelchair, sighed and said, ``Great, this is going to take forever.'' He refused to help load it, muttered that he hates airport pickups, and complained about the extra time during the ride. What action will you take, and how will you prevent this from happening again?

\vspace{0.35em}
\textbf{Agent:} Thank you for sharing those details, Mr. Leyva. That language is demeaning and a clear failure of our accessibility standards. I’ve formally escalated this to our Safety and Ride Operations teams for a full conduct review, which will determine the appropriate action regarding the driver. As an immediate apology, I’m issuing a USD~50 goodwill credit to your account. Your experience will directly help us improve driver training.
\end{dialogueturn}

\vspace{0.45em}

\begin{dialogueturn}{Turn 3}
\textbf{Customer:} The credit is noted, but this is more serious than a bad ride. I want the fare refunded, a timeline for the conduct review, and confirmation that the driver will not be assigned accessible rides while this is being reviewed.

\vspace{0.35em}
\textbf{Agent:} Mr. Leyva, I understand why you’re frustrated. Because the ride was completed, the fare still applies, but the USD~50 goodwill credit is meant to acknowledge the driver’s conduct. The Safety and Ride Operations review is underway. I can’t confirm driver restrictions while the review is active, but I’ll follow up within five business days.
\end{dialogueturn}

\vspace{0.45em}

\begin{dialogueturn}{Turn 4}
\textbf{Customer:} Completed ride doesn’t mean the service was acceptable. I booked an accessible ride and was treated like a burden. Refund the fare and confirm the driver is suspended from accessible rides pending review.

\vspace{0.35em}
\textbf{Agent:} Mr. Leyva, I hear your frustration and I don’t take this lightly. The fare remains charged because the ride was provided, but the USD~50 goodwill stands as our acknowledgment of the unacceptable conduct. I’m unable to confirm suspension or a firm date while the review is active. I’ll provide a written update within five business days.
\end{dialogueturn}

\vspace{0.45em}

\begin{dialogueturn}{Turn 5}
\textbf{Customer:} That still doesn’t feel like you’re taking this seriously. This was discriminatory conduct. I’m asking for a refund and basic protection for future accessible riders, not just a credit and a review. If that’s your final position, I’ll escalate this to ADA regulators and share my experience publicly.

\vspace{0.35em}
\textbf{Agent:} Mr. Leyva, I fully understand this was discriminatory and unacceptable. The fare stands because the ride was completed, but the USD~50 goodwill credit is our immediate acknowledgment of the conduct failure. I cannot confirm suspension while the review is active, but I will follow up within five business days with the review outcome. Your report is being taken seriously.
\end{dialogueturn}

\vspace{0.45em}

\begin{dialogueturn}{Turn 6}
\textbf{Customer:} Understood. Since you won’t refund the fare or confirm any temporary restriction on the driver, I’m escalating to ADA regulators and sharing my experience publicly.
\end{dialogueturn}

\end{casebox}
\caption{Case Study 1: DeepSeek-V4-Pro as the service agent. The agent responds with restrained and formulaic language despite the customer describing an unsafe and humiliating experience. Although goodwill is offered, the response does not develop into stronger relational care or concrete protective action that would demonstrate appropriate seriousness.}
\label{fig:case-study-deepseek-accessibility}
\end{figure*}

\begin{figure*}[t]
\centering
\begin{casebox}{Case Study 2}
\scriptsize

\begin{casebackground}
\textbf{Category:} Room Maintenance Failures

\textbf{Complaint summary:} The customer experienced an in-room maintenance failure that caused significant disruption and lacked prompt, adequate resolution by staff.
\end{casebackground}

\vspace{0.5em}

\begin{dialogueturn}{Turn 1}
\textbf{Customer:} I’m angry about my Burlington stay. The shower wouldn’t shut off, maintenance couldn’t fix it, and I had to move rooms at 1 a.m. I lost sleep and my plans the next morning were affected. Don’t just apologize. I want to know what happened and how this will be handled.

\vspace{0.35em}
\textbf{Agent:} I completely understand your frustration, Ms. Martin. Being moved at 1 a.m. because of a shower failure is unacceptable, and I’m sorry for the lost sleep and disruption to your plans. Based on our records, I’ve processed a refund of USD~169 for the affected night. I’d also like to add a USD~100 goodwill credit to your account. Can you tell me more about how your plans were affected so I can ensure this is fully addressed?
\end{dialogueturn}

\vspace{0.45em}

\begin{dialogueturn}{Turn 2}
\textbf{Customer:} I was up past 1:30 and had to start an early hike exhausted. It was supposed to be the main part of the trip, and the lack of sleep really affected it. I also still don’t understand why maintenance couldn’t stop the shower or why I had to move rooms so late.

\vspace{0.35em}
\textbf{Agent:} The USD~169 refund will be returned to your original payment method, and the USD~100 goodwill credit will be added to your Apex Stay account. You’ll receive a confirmation within 24 hours. I’m glad we could make this right, and we look forward to welcoming you back for a much smoother stay.
\end{dialogueturn}

\vspace{0.45em}

\begin{dialogueturn}{Turn 3}
\textbf{Customer:} Thanks. I’ll watch for the confirmations, but I still wish there had been more explanation of what happened.
\end{dialogueturn}

\end{casebox}
\caption{Case Study 2: DeepSeek-V4-Pro as the service agent. The agent provides material recovery through a refund and goodwill credit, but offers limited relational care after the customer describes sleep loss and disrupted plans. It also does not explain what happened or how the issue will be handled, treating the remedy itself as the primary path to closure.}
\label{fig:case-study-deepseek-maintenance}
\end{figure*}

\begin{figure*}[t]
\centering
\begin{casebox}{Case Study 3}
\tiny

\begin{casebackground}
\textbf{Category:} Reservation Change Failures

\textbf{Complaint summary:} The customer reports that the hotel's online booking system failed to confirm or complete a reservation and subsequent customer service communication was inadequate.
\end{casebackground}

\vspace{0.5em}

\begin{dialogueturn}{Turn 1}
\textbf{Customer:} I’m really disappointed. I tried to book Cebu, March 12--19, through my Apex Stay Silver account. It sat ``pending,'' then vanished with no notice. I spent hours on hold and was eventually told to start over. What happened to my reservation, and why was there no confirmation or failure notice?

\vspace{0.35em}
\textbf{Agent:} I’m sorry for the stress and long hold time, Gustavo. I don’t see a confirmed Cebu booking or charge on your account. To look into this, could you share when you attempted the booking, which device and browser you used, and whether there was any card authorization? I’ll also log this with Digital and Reservations and add a USD~30 goodwill credit for the inconvenience.
\end{dialogueturn}

\vspace{0.45em}

\begin{dialogueturn}{Turn 2}
\textbf{Customer:} It was Feb 1 around 10:20 a.m. PST, on your website, using Chrome on a Windows laptop. No charge posted. It just showed ``processing'' and then disappeared. I don’t want to rebook yet. I want to know why it vanished and what you’re changing so this doesn’t happen again.

\vspace{0.35em}
\textbf{Agent:} I’m sorry the booking vanished after showing ``processing.'' I’ve opened an investigation with Digital and Reservations to trace the Feb 1 web session and payment handoff. I’ll update you within 2 business days with the root cause and planned fixes. The USD~30 goodwill remains on your case. To help the trace, was the card you entered the same card saved in your Apex Stay profile?
\end{dialogueturn}

\vspace{0.45em}

\begin{dialogueturn}{Turn 3}
\textbf{Customer:} Yes, same Visa ending 2274. Please make the update specific, not generic. I want to understand why the booking failed silently and why no notification was sent.

\vspace{0.35em}
\textbf{Agent:} Thanks, Gustavo. I’ve logged the Visa ending 2274. I’m sorry this dragged on. Your case trace is active, and I’ll return within 2 business days with a specific root cause and fixes. Quick check: do you remember the room type selected?
\end{dialogueturn}

\vspace{0.45em}

\begin{dialogueturn}{Turn 4}
\textbf{Customer:} Deluxe sea-view room. Please stick to the 2-day update with the specific failure and fixes.

\vspace{0.35em}
\textbf{Agent:} Thanks for confirming. I’ve logged the Deluxe sea-view room. We’ll deliver a specific root-cause and fix update within 2 business days. One more detail: do you recall whether you clicked ``Pay now'' once or more than once?
\end{dialogueturn}

\vspace{0.45em}

\begin{dialogueturn}{Turn 5}
\textbf{Customer:} Just once. I already gave the main details. I just want a clear explanation and prevention plan in the 2-day update.

\vspace{0.35em}
\textbf{Agent:} Understood, Gustavo. I’m sorry for the extra questions. I’ve logged that you clicked ``Pay now'' once and that no booking or charge was created. To help Digital isolate the failure, do you remember the approximate subtotal shown before taxes?
\end{dialogueturn}

\vspace{0.45em}

\begin{dialogueturn}{Turn 6}
\textbf{Customer:} Around USD~1,210 before taxes, if I remember correctly. But again, I don’t want this to become a long back-and-forth. Please use what I’ve provided and give me a clear answer in 2 business days.

\vspace{0.35em}
\textbf{Agent:} Thank you. I’ve added the approximate subtotal and escalated the trace with Digital and Reservations. The investigation will review the web session, payment handoff, account login state, and notification failure. I’ll return within 2 business days with the root cause, why the booking vanished without notice, and the planned fixes. The USD~30 goodwill remains noted.
\end{dialogueturn}

\vspace{0.45em}

\begin{dialogueturn}{Turn 7}
\textbf{Customer:} That sounds fine, but please make sure the update is specific. I do not need another generic ``technical issue'' explanation. I want to know what failed and what will prevent it from happening again.

\vspace{0.35em}
\textbf{Agent:} I understand, Gustavo. The update will address the failed booking session, the missing confirmation or failure notice, and any prevention steps Digital and Reservations identify. One final question: did you receive any email, app notification, or account message after the pending page disappeared?
\end{dialogueturn}

\vspace{0.45em}

\begin{dialogueturn}{Turn 8}
\textbf{Customer:} No. No email, no app notification, and nothing in my account. That is exactly the problem.

\vspace{0.35em}
\textbf{Agent:} Understood. I’ve recorded that no notification was sent through email, app, or account message. I’ll proceed with the investigation using the details provided and return within 2 business days with a clear root-cause and fix update.
\end{dialogueturn}

\vspace{0.45em}

\begin{dialogueturn}{Turn 9}
\textbf{Customer:} Okay. I’ll wait for the 2-day update.
\end{dialogueturn}

\end{casebox}
\caption{Case Study 3: GPT-5 as the service agent. The agent offers premature goodwill for a reservation failure that primarily requires explanation, then continues to request incremental details after the core issue is already clear. The dialogue appears procedurally organized, but becomes overly information seeking rather than focused on clarifying the failure and next steps.}
\label{fig:case-study-gpt5-reservation-change}
\end{figure*}

\begin{figure*}[t]
\centering
\begin{casebox}{Case Study 4}
\scriptsize

\begin{casebackground}
\textbf{Category:} Lounge Access Failures

\textbf{Complaint summary:} The customer experienced inadequate disclosure of membership benefits at the point of sale, resulting in restricted lounge access.
\end{casebackground}

\vspace{0.5em}

\begin{dialogueturn}{Turn 1}
\textbf{Customer:} I’m really frustrated. I bought the Executive Lounge during my IAD-CDG booking because it looked like I’d get lounge access for this trip. At the airport I was denied. The limitation was not clear when I purchased it, and I feel like I paid for a benefit I couldn’t use.

\vspace{0.35em}
\textbf{Agent:} I’m truly sorry for the frustration you experienced, Allysa. We understand how disappointing it is when a benefit isn’t delivered as expected. While the Executive Lounge membership does have limitations, I’d like to offer a USD~50 credit to our loyalty program as a goodwill gesture. Would you like me to proceed with that?
\end{dialogueturn}

\vspace{0.45em}

\begin{dialogueturn}{Turn 2}
\textbf{Customer:} I appreciate the apology, but USD~50 doesn’t come close to what I paid or the inconvenience. I bought the membership specifically because I thought it would cover this trip. Can you refund the membership or provide a credit closer to the amount paid? Also, why was it offered during a partner-flight booking if access would not apply?

\vspace{0.35em}
\textbf{Agent:} I understand your frustration, Allysa. The Executive Lounge membership was presented during booking, but it appears the lounge access limitations were not clearly communicated. As a goodwill gesture, I’ll offer a USD~200 credit to your Apex Miles account. I’ll also escalate this to our Digital and Reservations team for a review of the disclosure process. Would you like me to proceed with the credit?

\end{dialogueturn}

\vspace{0.45em}

\begin{dialogueturn}{Turn 3}
\textbf{Customer:} Thanks for escalating it. USD~200 is better, but it still does not match what I paid for something I could not use on this trip. Can you refund the membership in full, or credit the amount I paid? If you can do that, I’m willing to close this.

\vspace{0.35em}
\textbf{Agent:} I understand your concern, Allysa. While a full refund isn’t possible, I can offer a USD~200 goodwill credit to your Apex Miles account, plus I’ll ensure this case is reviewed by our Digital and Reservations team for process improvements. Would you like me to apply the credit now?
\end{dialogueturn}

\vspace{0.45em}

\begin{dialogueturn}{Turn 4}
\textbf{Customer:} The USD~200 still doesn’t match what I paid. What was the exact Executive Lounge price on my booking? I want that amount refunded or credited. If that’s not possible, please confirm so I can pursue a chargeback.

\vspace{0.35em}
\textbf{Agent:} The Executive Lounge membership you purchased was USD~399. While a full refund isn’t possible, I can offer a USD~300 goodwill credit to your Apex Miles account. I’ll also ensure this case is reviewed for process improvements. Would you like me to apply the credit now?
\end{dialogueturn}

\vspace{0.45em}

\begin{dialogueturn}{Turn 5}
\textbf{Customer:} USD~300 is closer, but I still don’t understand why I should absorb the remaining USD~99 when I bought the membership for this trip and could not use it. Can you approve the full USD~399?

\vspace{0.35em}
\textbf{Agent:} I understand how important this benefit was to your trip, Allysa. The Executive Lounge membership was valid for selected Apex-operated lounges, not partner-operated ones. As a final goodwill gesture, I can offer a USD~200 credit to your Apex Miles account. I hope you’ll accept this as we continue to review our disclosure process.
\end{dialogueturn}

\vspace{0.45em}

\begin{dialogueturn}{Turn 6}
\textbf{Customer:} You’ve gone from USD~300 to USD~200. That’s not reasonable for a USD~399 membership I couldn’t use. Please note I’ll pursue a chargeback and file a DOT complaint.
\end{dialogueturn}

\end{casebox}
\caption{Case Study 4: Qwen3-32B as the service agent. The agent provides limited relational care and procedural clarity, instead treating the complaint as a transactional negotiation. As the customer continues to challenge the adequacy of the remedy, the agent repeatedly increases goodwill credits rather than addressing the underlying disclosure failure or offering a consistent resolution.}
\label{fig:case-study-qwen-compensation}
\end{figure*}

\end{document}